\documentclass{article}
\usepackage{arxiv_preprint,times}

\usepackage{xcolor}
\usepackage{hyperref}
\usepackage{url}
\usepackage{graphicx}
\usepackage{amsmath,amssymb,bm}
\usepackage{booktabs}
\usepackage{tabularx}
\usepackage{cleveref}

\definecolor{linkcolor}{RGB}{0,70,140}
\hypersetup{
  colorlinks=true,
  linkcolor=linkcolor,
  citecolor=linkcolor,
  urlcolor=linkcolor,
  pdftitle={Dataset-Level Regularization for Joint-Embedding Self-Supervised Learning},
  pdfauthor={Konstantinos Almpanakis, Anna Kreshuk}
}

\newcommand{\dataset}{\mathcal D}
\newcommand{\regularizer}{\mathcal R}
\newcommand{\linv}{\mathcal L_{\mathrm{align}}}
\newcommand{\lanti}{\mathcal L_{\mathrm{ac}}}
\newcommand{\quantreg}{\textsc{SIGReg-OOB}}
\newcommand{\contooob}{\textsc{Contrastive-OOB}}
\newcommand{\sigreg}{\textsc{SIGReg}}

\newcommand{\sphere}{\mathbb S}

\title{Dataset-Level Regularization for\\Joint-Embedding Self-Supervised Learning}

\author{%
  Konstantinos Almpanakis\thanks{Corresponding author.}
  \And
  Anna Kreshuk
  \AND
  European Molecular Biology Laboratory (EMBL)\\
  Heidelberg, Germany\\[0.5ex]
  \small\texttt{konstantinos.almpanakis@embl.de}\quad
  \small\texttt{anna.kreshuk@embl.de}
}

\begin{document}

\maketitle

\begin{abstract}
Self-supervised joint embedding learning aligns representations of augmented views while regularizing their distribution to prevent collapse. Regularization aims to shape the overall representation distribution, but most joint embedding methods restrict its computation to the active minibatch. We propose out-of-batch regularization,
which maintains a dataset-level state in CPU memory and uses globally constructed per-instance targets during minibatch training. Evaluated on natural images and biomedical datasets, these anchors approximate the full-dataset regularization gradient more faithfully than conventional minibatch estimation at small batch sizes. Applications to contrastive (SimCLR) and non-contrastive (SIGReg) approaches show consistent improvement over the corresponding in-batch objectives for image classification, minority-class performance under controlled imbalance,
instance retrieval, and continued self-supervised pretraining, with the largest gains at small batch sizes. These results suggest that dataset-level regularization can be decoupled from minibatch optimization, making batch size an optimization choice that does not affect the geometry of the regularizer.

\end{abstract}

\section{Introduction}
\label{sec:introduction}


Joint-embedding architectures (JEAs) \citep{chen2020simple,grill2020bootstrap,caron2021emerging,bardes2021vicreg} have been the dominant paradigm for visual self-supervised representation learning. Applications range from 3D medical image analysis
\citep{tang2022self} to 3D scene reconstruction \citep{wang2025vggt} and vision--language alignment \citep{radford2021learning}. They use augmentations to create a pretraining objective based on view invariance. Anti-collapse regularization complements this objective to prevent trivial solutions \citep{chen2021exploring}. While different regularization mechanisms have been proposed over the years, most of them act on the embeddings of the current minibatch (or a few recent batches with EMA) that might be too small to represent the dataset under memory or optimization constraints.

The bottleneck is particularly relevant for high-dimensional inputs, including high-resolution images and 3D volumes, where each sample requires more activation memory. Such applications are thus often limited to small-batch training. The problem worsens for imbalanced data, where rare classes may contribute only a few examples to each batch \citep{jing2021understanding}. The simple solution of using a larger batch is constrained by hardware \citep{gao2021scaling}, and even when a large batch is possible, it may not be ideal for optimization. On small datasets, using a large fraction of the training samples per batch leaves few parameter updates per epoch. \citet{masters2018revisiting} also report that larger batches can
require more careful choice of the learning rate and training duration. The batch size needed to estimate a population regularizer can therefore differ from the best batch size for optimization.

\begin{figure*}[h]
  \centering
  \includegraphics[width=0.96\textwidth]{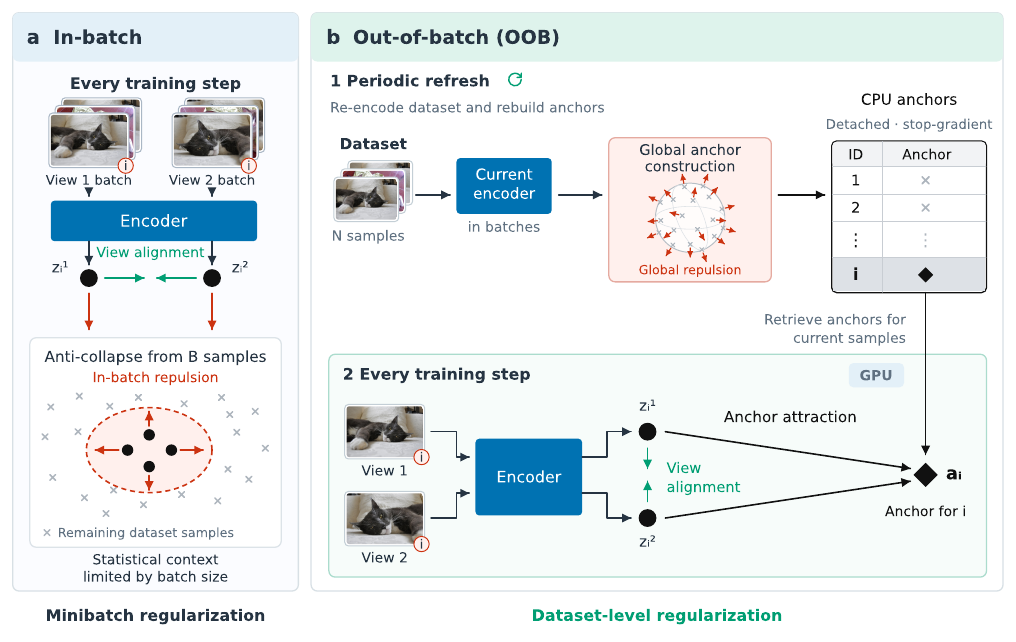}
  \caption{\textbf{Minibatch and OOB regularization.}
  (a) In-batch methods estimate the anti-collapse term from active samples at each step. (b) OOB regularization periodically encodes the
  dataset, constructs detached per-instance anchors, and
  retrieves the anchors indexed by the online minibatch.}
  \label{fig:overview}
\end{figure*}

We propose a framework that moves the statistical context of SSL regularizers from the active batch to the dataset-level state. We maintain a detached, dataset-indexed state in CPU memory. A periodic refresh encodes the training set and constructs globally optimized targets, while online steps retrieve only the current samples’ targets (Fig.~\ref{fig:overview}). Thus, the support of the anti-collapse component is no longer limited by the batch size or GPU capacity.


Our main contributions are:
\begin{itemize}
  \item We introduce OOB regularization, where offline, dataset-level
  targets act as per-instance anchors during online training, and derive
  contrastive (\contooob{}) and non-contrastive (\quantreg{}) instances of this construction.
  \item We formulate anti-collapse as estimation of a finite population
  gradient and measure minibatch error and cache staleness on 19 real datasets.
  The analysis also separates gradient-estimation error from population
  reweighting under class imbalance.
  \item Across natural images and 2D/3D biomedical data, the OOB variants
  yield positive mean gains over their matched in-batch objectives in image
  classification evaluated by kNN and linear probes. The gains are largest at
  small batch sizes and extend to minority-class recall, instance retrieval,
  and continued pretraining.
\end{itemize}

\section{Related Work}
\label{sec:related}

\textbf{Collapse prevention in joint embedding training.}
Joint-embedding methods must preserve informative variation while aligning
views. Contrastive objectives achieve this through negative-sample repulsion
\citep{oord2018representation,chen2020simple}, linking alignment to
hyperspherical uniformity for normalized representations
\citep{wang2020understanding}. To avoid explicit negatives, Barlow Twins and
VICReg constrain correlations and moments
\citep{zbontar2021barlow,bardes2021vicreg}, while SIGReg matches projected
distributions to a Gaussian \citep{balestriero2025lejepa}. Other approaches
prevent collapse through balanced assignments in SwAV
\citep{caron2020unsupervised} or asymmetry, teachers, and stop-gradient
in BYOL, DINO, and SimSiam
\citep{grill2020bootstrap,caron2021emerging,chen2021exploring}.
These approaches establish how to prevent collapse, but they have to assume that the active batch is representative of the dataset.  

\textbf{Cross-batch state and global objectives.}
Reusing representations across updates expands batch context. MoCo's momentum-encoder queue \citep{he2020momentum,chen2020improved},
instance discrimination's memory bank \citep{wu2018unsupervised}, and
cross-batch memory \citep{wang2020cross} expand the pool of negative samples. NNCLR uses stored embeddings to
find positive neighbors \citep{dwibedi2021little}.
\citet{chen2022we} argues that minibatch contrastive gradients can be biased, motivating global-objective
optimization through per-image moving averages in SogCLR
\citep{yuan2022provable} and MCMC negative sampling in EMC$^2$
\citep{yau2024emc}. Decoupled Contrastive Learning reduces batch sensitivity
by removing positives from the denominator \citep{yeh2022decoupled}. OOB instead performs population
regularization periodically and retrieves per-instance targets for online
attraction, avoiding cross-instance comparisons in the online regularizer.

\textbf{Assigned targets and dataset-indexed representations.}
Constructing targets separately from encoder updates also provides broader context. Noise-vector assignment
\citep{bojanowski2017unsupervised}, dataset-feature clustering
\citep{caron2018deep}, and balanced self-labeling through optimal transport
\citep{asano2019self} organize representation learning around assigned targets.
Our anchors instead retain per-instance correspondence and are reconstructed
from current representations under a specified continuous geometry. DIET
also uses sample identity, predicting datum indices as classes
\citep{balestriero2023unsupervised}. IConE, the conceptually closest recent approach \citep{almpanakis2026icone}, separates collapse prevention from encoder alignment through a jointly optimized \(N\times d\) table of auxiliary instance embeddings. Its regularization maintains a persistent global parameter space during training and evaluates an \(N\times N\) similarity matrix each step. OOB instead periodically constructs detached per-instance targets from a chosen population-level regularizer, stores these targets in CPU memory, and retrieves only those required by the active minibatch. This decouples online training from the global computation while supporting different regularizers.

\section{Method}
\label{sec:method}

To give small-batch training access to dataset-wide regularization, we
construct anti-collapse anchors from the full dataset and use them to update
the encoder. We first formulate this
separation, instantiate it with spherical (contrastive) and Gaussian (non-contrastive) anchors, and then
define a full-dataset gradient reference for comparing minibatch estimation
with cached regularization.

\subsection{Alignment and population regularization}

We use mutual information as an organizing view of alignment and anti-collapse, without assuming that joint-embedding methods directly optimize it.
For representations $Z_1$ and $Z_2$ of related views, the mutual information (MI)
identity \citep{shannon1948mathematical} gives
\begin{equation}
  I(Z_1;Z_2)=H(Z_1)-H(Z_1\mid Z_2).
  \label{eq:mi}
\end{equation}
This identity motivates two complementary roles in JEA losses: alignment reduces
uncertainty between views, while anti-collapse keeps the marginal representation informative. Representation quality nevertheless depends on more than MI \citep{tschannen2019mutual}.

Let $\dataset=\{x_i\}_{i=1}^{N}$ be a dataset of unlabeled
inputs, and let $f_\theta$ map an input to its representation. For a sample
$x$ drawn uniformly from $\dataset$, let $t_1,t_2$ be independently sampled
transformations from an augmentation family $\mathcal T$, producing views
$t_1(x)$ and $t_2(x)$. A broad JE-SSL objective is
\begin{equation}
  \mathcal L_{\mathrm{SSL\text{-}JE}}(\theta)
  =\underbrace{\mathbb E_{x,t_1,t_2}\!\left[
  \linv(f_\theta(t_1(x)),f_\theta(t_2(x)))\right]}_{\text{Alignment}}
  +\underbrace{\lambda\regularizer(P_\theta)}_{\text{Anti-collapse}},
  \label{eq:population-loss}
\end{equation}
where $P_\theta$ is the distribution of $f_\theta(t(x))$ induced by sampling
inputs and augmentations.


In practice, a regularizer sees a finite set of representations. Writing
$z_i=f_\theta(t(x_i))$, the empirical distributions of an active minibatch
$\mathcal B$ of size $B$ and the full training set are
\begin{equation}
  \widehat P_B=\frac{1}{B}\sum_{i\in\mathcal B}\delta_{z_i},
  \qquad
  \widehat P_N=\frac{1}{N}\sum_{i=1}^{N}\delta_{z_i}.
  \label{eq:empirical-populations}
\end{equation}
Here $\delta_{z_i}$ is a unit point mass at \(z_i\), giving each included representation equal probability.
Standard in-batch training evaluates $\regularizer(\widehat P_B)$ in place
of the population term in \cref{eq:population-loss}. Our aim is to supply
online updates with the broader context of $\regularizer(\widehat P_N)$
without recomputing it through the encoder at every step. We therefore
construct dataset-level anchors periodically and use a separable
sample-to-anchor loss between refreshes.

\subsection{A refresh-and-train interface}

Let
$z_i^{(v)}$ denote the representation of augmented view $v$ of
sample $i$. Our OOB methods optimize
\begin{equation}
  \mathcal L_{\mathrm{OOB}}
  =\lambda_{\mathrm{align}}\linv
  +\lambda_{\mathrm{ac}}\lanti^{\mathrm{OOB}},
  \label{eq:oob-loss}
\end{equation}
where $\linv$ maximizes view invariance and $\lanti^{\mathrm{OOB}}$ attracts each representation
to its dataset-indexed anchor. To implement $\linv$, we maximize cosine similarity between embeddings produced by different views of the same instance. To compute $\lanti^{\mathrm{OOB}}$, at refresh $r$, the current encoder processes one view of every
training sample. A population operation (\cref{sec:quantreg,sec:coob}) converts outputs into detached
anchors $A^{(r)}=\{a_i^{(r)}\}_{i=1}^{N}$, which are stored in CPU memory. For
an online minibatch $\mathcal B$, only its indexed rows are moved to the
GPU (\Cref{tab:oob-algorithm}):
\begin{equation}
  \lanti^{\mathrm{OOB}}
  =\frac{1}{B V}\sum_{i\in\mathcal B}\sum_{v=1}^{V}
  d_{\mathcal G}\!\left(h_r(z_i^{(v)}),a_i^{(r)}\right).
  \label{eq:oob-anchor-loss}
\end{equation}
Here $d_{\mathcal G}$ measures discrepancy from the anchor using MSE
across projections for Gaussian anchors, or cosine distance for $L_2$-normalized spherical anchors; $h_r$ applies refresh-fixed projections (Gaussian anchors) or $L_2$ normalization (spherical anchors). Gradients pass through $z_i$, but not through the anchors, and
the online anti-collapse term requires no relationship among samples in
$\mathcal B$.

\begin{table}[h]
  \centering
  \caption{OOB training separates global target construction from online
  parameter updates.}
  \label{tab:oob-algorithm}
  \footnotesize
  \setlength{\tabcolsep}{3pt}
  \begin{tabularx}{\columnwidth}{@{}rX@{}}
    \toprule
    & \textbf{Refresh phase} \\
    1 & Run inference in minibatches over the entire dataset, collecting one view representation per sample. \\
    2 & After the full inference pass, construct $a_i^{(r)}$ jointly from all $N$ representations. \\
    3 & Detach $A^{(r)}$ and store it by sample index in CPU memory. \\
    \midrule
    & \textbf{Online phase, repeated until the next refresh} \\
    4 & Sample $\mathcal B$ and generate $V$ augmented views per sample. \\
    5 & Retrieve $\{a_i^{(r)}:i\in\mathcal B\}$ and evaluate
        \cref{eq:oob-loss}. \\
    6 & Update encoder and projector parameters; keep anchors fixed. \\
    \bottomrule
  \end{tabularx}
\end{table}

\subsection{\quantreg{}: projected Gaussian anchors}
\label{sec:quantreg}

\quantreg{} constructs a Gaussian target through one-dimensional rank
transport assignment to a fixed
Gaussian quantile grid \citep{gabriel2019computational}. At refresh, choose unit directions $v_1,\ldots,v_M$. For
$s_{im}=v_m^\top z_i$, sort all $N$ samples and assign the rank-$k$ sample to
\begin{equation}
  \gamma_k=\Phi^{-1}\!\left(\frac{k+1/2}{N}\right),
  \qquad k=0,\ldots,N-1.
  \label{eq:quantiles}
\end{equation}
If $r_{im}$ is the rank of sample $i$ along
$m$, its cached coordinate is $q_{im}=\gamma_{r_{im}}$. The online term is
\begin{equation}
  \lanti^{\text{SIGReg-OOB}}
  =\frac{1}{BVM}\sum_{i\in\mathcal B}\sum_{v=1}^{V}\sum_{m=1}^{M}
  \left(v_m^\top z_i^{(v)}-q_{im}\right)^2.
  \label{eq:quantreg-loss}
\end{equation}
Refresh inference uses GPU, while the $M$ sorts and the
$N\times M$ target table are computed on CPU. We resample the
projection directions at each refresh. Our \sigreg{} baseline uses the same sorted-quantile targets computed in the batch, rather than the Epps-Pulley statistic \citep{balestriero2025lejepa}, so we keep the same regularizer. We ablate \quantreg{} hyperparameters in \cref{app:hparam-ablations}.



\subsection{\contooob{}: spherical anchors}
\label{sec:coob}

For contrastive learning, anchors lie on the unit sphere. Starting from the
normalized refresh representations $\bar z_i^{(r)}$, we define the population
target by
\begin{align}
  \mathcal U(A)&=\log\!\left[
  \frac{1}{N(N-1)}\sum_{i\ne j}
  e^{-\tau_u\lVert a_i-a_j\rVert_2^2}\right], \\
  A^{(r)}&\approx\arg\min_{A\in(\sphere^{d-1})^N}
  \mathcal U(A)+\frac{\lambda_f}{N}\sum_i
  \left(1-a_i^\top\bar z_i^{(r)}\right).
  \label{eq:spherical-anchor-objective}
\end{align}
The first term favors a globally spread spherical population
\citep{wang2020understanding}; the second preserves anchor–sample correspondence. With
$u_i^{(v)}=z_i^{(v)}/\lVert z_i^{(v)}\rVert_2$, online training uses
\begin{equation}
  \lanti^{\text{Contrastive-OOB}}
  =\frac{1}{BV}\sum_{i\in\mathcal B}\sum_{v=1}^{V}
  \left(1-{u_i^{(v)}}^\top a_i^{(r)}\right).
  \label{eq:coob-loss}
\end{equation}

We optimize Eq.~\ref{eq:spherical-anchor-objective} stochastically, avoiding
all $N(N-1)$ pairs. Each of 20 steps evaluates all pairs among up to 256 sampled anchors. The first refresh initializes from
the source plus small Gaussian noise. SimCLR obtains spherical spreading from minibatch negatives, while
MoCoV2 extends its negative support with a momentum queue. \contooob{} instead
performs repulsion during anchor construction and uses only sample-to-anchor
attraction online, without in-batch negatives or a momentum encoder. This
preserves the intended spherical spreading geometry that contrastive learning imposes. For specific training settings, see \cref{app:method-hyperparameters}.

\subsection{Comparing minibatch and cached gradients}
\label{sec:population}

Having defined the training methods, we compare a minibatch estimate and a cached estimate against the same full-dataset gradient at the current encoder
state. 
Differential entropy is a property of the underlying distribution, but
training provides only a finite set of encoder outputs. Practical methods optimize
finite-sample surrogates that spread or decorrelate these
outputs. Using the full empirical distribution in \cref{eq:empirical-populations},
the reference for a chosen surrogate is its full empirical gradient
\begin{equation}
  g_i^\star=\nabla_{z_i}\regularizer(\widehat P_N).
  \label{eq:oracle}
\end{equation}
We call $g_i^\star$ the oracle gradient. It is the representation-space
direction supplied by the chosen regularization when that surrogate
sees the complete empirical population. Minibatch gradient bias for contrastive objectives has been established in prior work
\citep{chen2022we,yuan2022provable}, but here we compare representation-space gradients for a fixed finite-population regularizer.
The standard minibatch estimator substitutes the active empirical distribution
$\widehat P_B$:
\begin{equation}
  \widehat g_{i,B}=\nabla_{z_i}\regularizer(\widehat P_B).
  \label{eq:minibatch-gradient}
\end{equation}
An OOB estimator instead constructs dataset state
$\mathcal C_{t_0}=\{c_i\}_{i=1}^{N}$ at refresh step $t_0$ and later uses
\begin{equation}
  \widetilde g_{i,t}=\nabla_{z_i}\lanti(z_i,c_i),\qquad t\ge t_0.
  \label{eq:cached-gradient}
\end{equation}
We measure directional agreement of cached and minibatch approximations with the oracle using cosine similarity and scale-sensitive error using \(\operatorname{rRMSE}\).

\textbf{Computing the reference and its estimators.}
All three estimators differentiate the same error between projected
representations and Gaussian quantile targets (SIGReg, \quantreg{}). The oracle ranks the full current dataset, the minibatch estimator
ranks each probe within sampled batches of size $B$,
and the cached estimator reuses full-dataset assignments from refresh step
$t_0$. At cache age $s$, we recompute the oracle using the current encoder at
$t_0+s$, so the comparison measures staleness against the current population. Experimental results are in \cref{sec:oracle-results}.

\textbf{Separating estimation from population weighting.}
Class imbalance also leads to the desired representation prior not matching the data's class proportions \citep{assran2022hidden}.
If class $c$ has $N_c$ samples,
the empirical population weights it by $N_c/N$. To separate estimator error
from the weighting, we sample another population in a label-aware manner for evaluation:
\begin{equation}
  \widehat P_{\mathrm{bal}}=
  \frac{1}{C}\sum_{c=1}^{C}\frac{1}{N_c}
  \sum_{i:y_i=c}\delta_{z_i}.
  \label{eq:balanced-population}
\end{equation}
Comparing its gradient with the empirical-population
gradient separates a change in the objective's weighting from an error in estimating a fixed objective. Gradient comparisons and downstream imbalance results appear in \cref{sec:imbalance-results}.

\section{Results}
\label{sec:experiments}

We examine gradient-estimation accuracy (\cref{sec:oracle-results}), frozen
classification performance across batch sizes (\cref{sec:classification-results}),
population weighting and minority-class recall under imbalance
(\cref{sec:imbalance-results}), instance retrieval
(\cref{sec:retrieval-results}), adaptation of pretrained encoders
(\cref{sec:continued-results}), and performance under GPU-memory constraints
(\cref{sec:memory-results}).

\subsection{Datasets}
\label{sec:datasets}

We experiment on natural and scientific/biomedical images: CIFAR-10/100 \citep{krizhevsky2009learning},
DTD \citep{cimpoi2014describing}, STL-10 \citep{coates2011analysis}, ImageNet-100, Imagenette, and Imagewoof
\citep{deng2009imagenet}; BBBC010
\citep{ljosa2012annotated}, twelve 2D and six 3D
MedMNIST datasets \citep{yang2023medmnist}, and a private plankton dataset imaged with widefield microscopy. Here we present aggregate results, see \cref{app:classification} for complete dataset-level results and size sweeps. 


\subsection{Oracle gradient analysis}
\label{sec:oracle-results}

\begin{figure}[h]
  \centering
  \includegraphics[width=\textwidth]{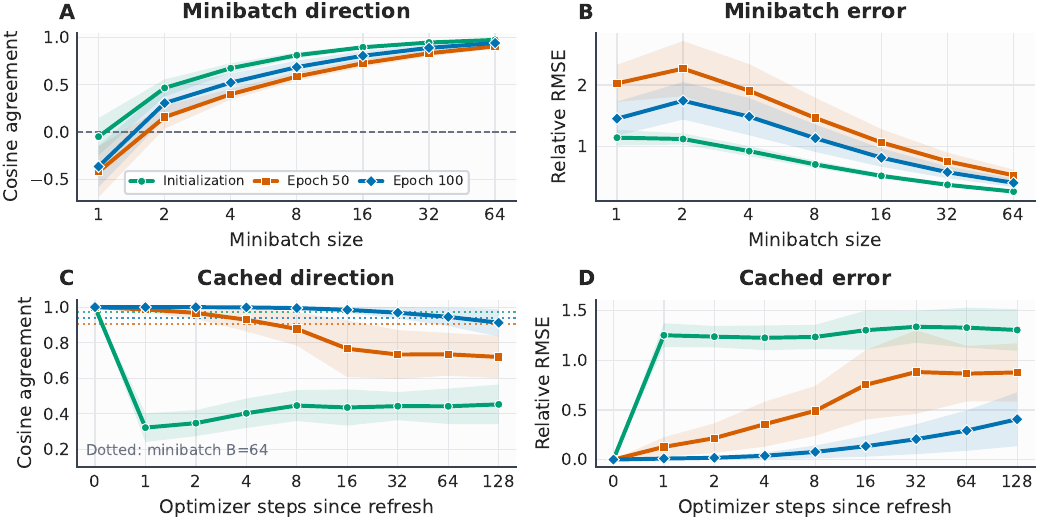}
  \caption{Minibatch and cached gradient agreement with the full-dataset reference.}
  \label{fig:oracle}
\end{figure}

Using the fixed objective from \cref{sec:population}, we compare
minibatch sampling error with cache staleness at training
checkpoints. \Cref{fig:oracle}A--B shows that limited batch context can change the
direction of regularization and can produce gradients that oppose the full-dataset objective. As batch size gets bigger, its regularization gradient agrees more with the oracle. A fresh cache recovers the oracle exactly, but its agreement deteriorates as
the encoder changes (\Cref{fig:oracle}C--D). This deterioration is slower at later checkpoints, where cached
gradients remain informative across more online updates. The results confirm that minibatch and
cached estimators face different limitations: incomplete population context
versus temporal staleness. When training with small batches, periodic dataset-level regularization is clearly beneficial, while the update schedule should depend on the training stage.

\subsection{Image classification across batch sizes}
\label{sec:classification-results}

\textbf{Evaluation protocol.} We compare each OOB method with its corresponding baseline: \quantreg{}
with \sigreg{}, and \contooob{} with SimCLR and MoCoV2. Within a comparison,
we keep the data split, augmentations, number of views, backbone, projector,
optimizer, learning rate schedule, and evaluation fixed. The 2D experiments use a
ViT backbone \citep{dosovitskiy2020image}; 3D experiments use a 3D ResNet \citep{he2016deep} and training lasts 100 epochs. Frozen representations
are evaluated by k-NN and linear probe
\citep{wu2018unsupervised,chen2020simple}.
We use three independent seeds. In-batch baselines are omitted
at $B=1$, because anti-collapse regularization based on a single-sample batch is either undefined or statistically uninformative. 
OOB methods encode dataset-level state and remain well defined. Training details are in \cref{app:training-configuration}.

\begin{figure}[h]
  \centering
  \includegraphics[width=\textwidth]{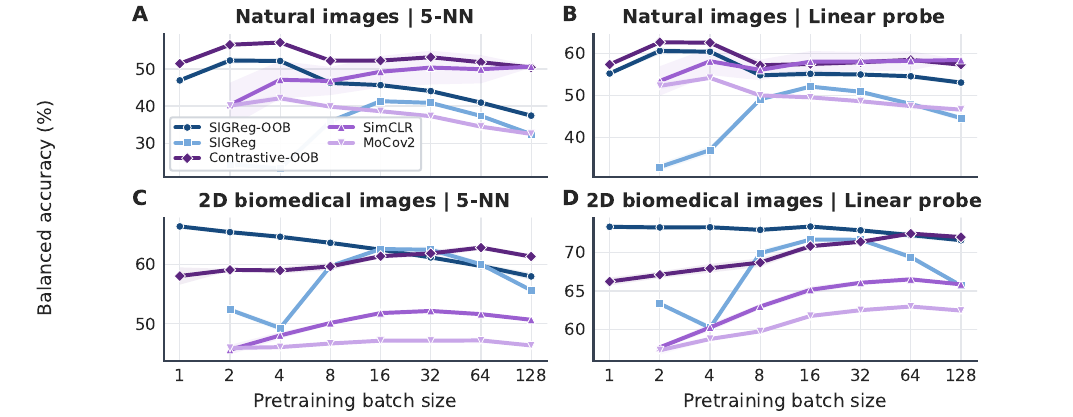}
  \caption{Image classification results using linear probes and 5-NN.}
  \label{fig:classification}
\end{figure}

\textbf{Results.} The low-batch regime shows the clearest separation (\Cref{fig:classification}).
Averaged over small batch sizes, \quantreg{} improves 5-NN/linear probe performance over
\sigreg{} by $14.17/12.89$pp on natural images, while \contooob{} improves over
SimCLR by $10.83/6.58$pp on biomedical data. On 3D volumes, both OOB
methods retain positive mean gains, showing that the effect is not tied to a
2D encoder. Complete paired aggregates are in
\Cref{tab:classification-gains}. \Cref{app:classification} adds comparisons with other baselines like IConE, BYOL, DINO, SwaV and VICReg. \Cref{app:rank} contains representation metrics like RankMe and effective-rank \citep{garrido2023rankme,roy2007effective}.

\begin{table}[h]
  \centering
  \caption{Classification balanced accuracy (\%). Parentheses show OOB gains in percentage points.}
  \label{tab:classification-gains}
  \small
  \definecolor{positivechange}{HTML}{178A50}
\definecolor{negativechange}{HTML}{C43D3D}
\setlength{\tabcolsep}{2.5pt}
\begin{tabular*}{\textwidth}{@{\extracolsep{\fill}}l*{8}{r}@{}}
\toprule
 & \multicolumn{2}{c}{\sigreg{}} & \multicolumn{2}{c}{\textsc{SIGReg-OOB} (Ours)} & \multicolumn{2}{c}{SimCLR} & \multicolumn{2}{c}{\contooob{} (Ours)} \\
\cmidrule(lr){2-3}\cmidrule(lr){4-5}\cmidrule(lr){6-7}\cmidrule(l){8-9}
Benchmark & 5-NN & Linear & 5-NN & Linear & 5-NN & Linear & 5-NN & Linear \\
\midrule
Natural images & 34.63 & 45.55 & 48.80 (\textcolor{positivechange}{+14.17}) & 58.44 (\textcolor{positivechange}{+12.89}) & 50.21 & 59.35 & 52.50 (\textcolor{positivechange}{+2.29}) & 59.49 (\textcolor{positivechange}{+0.14}) \\
2D biomedical & 57.45 & 67.44 & 62.21 (\textcolor{positivechange}{+4.76}) & 72.84 (\textcolor{positivechange}{+5.40}) & 50.00 & 63.55 & 60.83 (\textcolor{positivechange}{+10.83}) & 70.13 (\textcolor{positivechange}{+6.58}) \\
3D biomedical & 51.39 & 55.89 & 55.61 (\textcolor{positivechange}{+4.22}) & 63.95 (\textcolor{positivechange}{+8.06}) & 51.76 & 57.52 & 55.47 (\textcolor{positivechange}{+3.71}) & 61.78 (\textcolor{positivechange}{+4.26}) \\
\bottomrule
\end{tabular*}

\end{table}

\subsection{Population weighting and minority classes under imbalance}
\label{sec:imbalance-results}

\begin{figure}[htbp]
  \centering
  \includegraphics[width=0.93\textwidth]{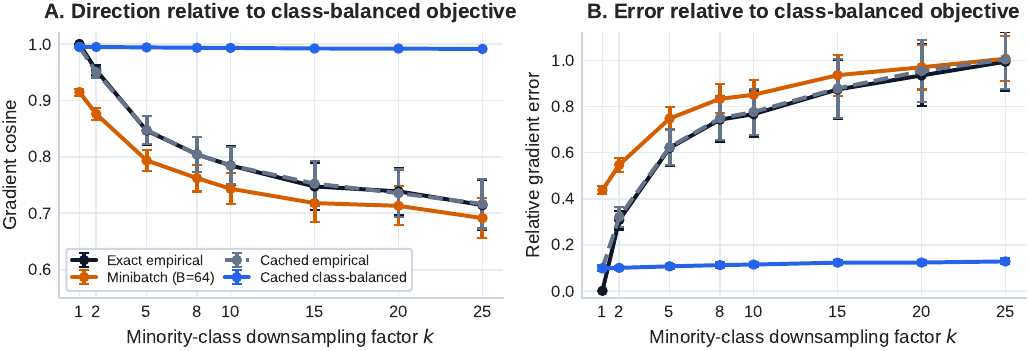}
  \caption{Empirical and class-balanced gradient agreement under class imbalance.}
  \label{fig:minority-oracle}
\end{figure}

\textbf{Oracle diagnostic.}
Class-balanced reference
\cref{eq:balanced-population} is used to distinguish estimator error from the
choice of population weighting.
The exact empirical gradient for
minority probes diverges from the class-balanced gradient as minority classes
are downsampled (\cref{fig:minority-oracle}). A cached empirical estimator tracks the empirical objective closely and a cache built with class-balanced targets tracks the class-balanced oracle. OOB state solves the estimation problem, but not class imbalance. When class information is available, population weighting can be used as a hyperparameter.

\begin{figure}[h]
  \centering
  \includegraphics[width=0.92\textwidth]{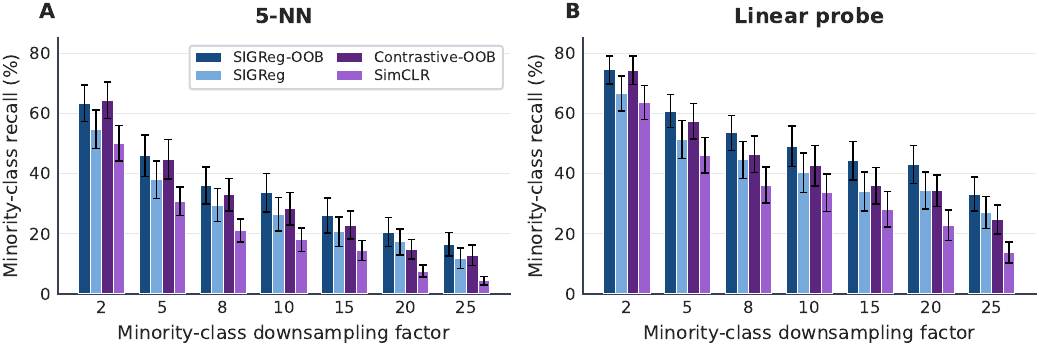}
  \caption{Minority-class recall under controlled class imbalance.}
  \label{fig:unbalanced}
\end{figure}

\textbf{Downstream evaluation.}
Having separated these effects, we test whether OOB anchors improve
minority-class representations without using labels during training. We create imbalanced MedMNIST variants by downsampling some classes by
$k$. We report macro-average recall over downsampled classes using the same held-out split for matched methods. Across different batch sizes, \quantreg{} gains $6.24$ points in 5-NN minority recall
and $8.6$ points in linear minority recall over \sigreg{}. \contooob{} gains
$10.69$ and $10.28$ points over SimCLR. The gains remain positive across the
downsampling sweep (\Cref{fig:unbalanced}). These downstream
results complement \Cref{fig:minority-oracle}: an empirical OOB cache does not
rebalance the population, but it gives each rare sample a global context that
is often absent from its minibatch. Full 2D/3D per-dataset results, worst-class recall, and
minority--majority gaps are in \cref{app:imbalance}.

\subsection{Instance retrieval}
\label{sec:retrieval-results}

\begin{figure}[h]
  \centering
  \includegraphics[width=\textwidth]{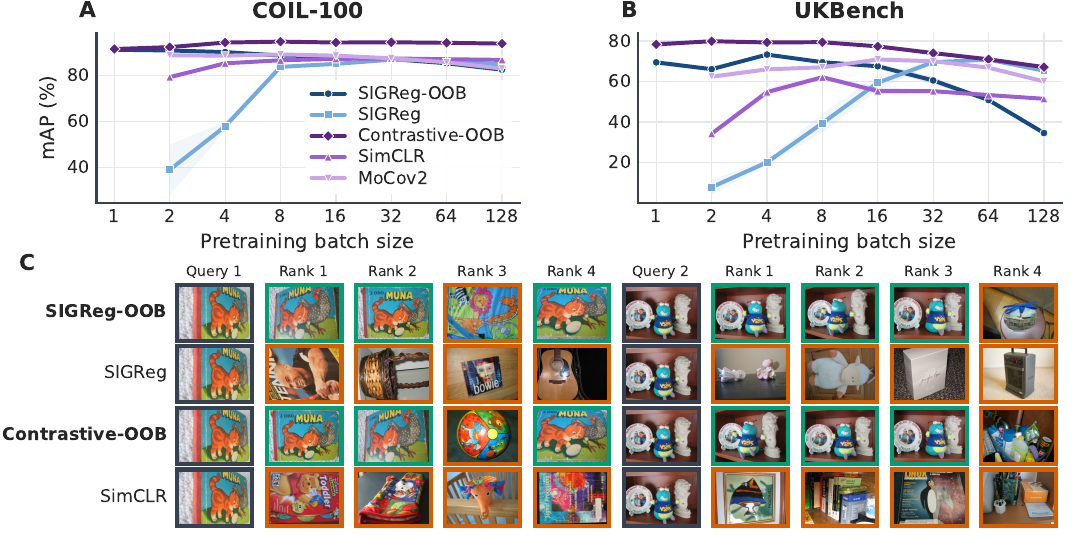}
  \caption{Instance-retrieval mAP and example nearest neighbors.}
  \label{fig:retrieval}
\end{figure}

We next test whether the learned space preserves instance neighborhoods in addition to the class boundaries. After pretraining, we retrieve with cosine similarity between frozen features. We report mean average
precision (mAP) over three seeds and show the four nearest neighbors of the same query by each method (\cref{fig:retrieval}). Across our experiments, \quantreg{} improves mAP over
\sigreg{} by $13.1$ points. \contooob{} improves over SimCLR/MoCoV2 by $13.7/7.6$ points, at a given batch size. Recall@$1/5/10$ and more retrieval evaluations are in \cref{app:retrieval}.

\subsection{Continued self-supervised pretraining}
\label{sec:continued-results}

Continued SSL pretraining adapts an existing model to a
target domain \citep{reed2022self,rodas2025diet}.
Small batches can make a larger pretrained encoder feasible during
domain adaptation. We train DINOv2 \citep{oquab2023dinov2} of several sizes
on our datasets, without labels.

\begin{table}[htbp]
  \centering
  \caption{Continued pretraining of DINOv2 and 5-NN of adapted models relative to original.}
  \label{tab:continued}
  \scriptsize
  \setlength{\tabcolsep}{7pt}
  \renewcommand{\arraystretch}{0.92}
\setlength{\tabcolsep}{2.0pt}
\resizebox{\textwidth}{!}{%
\begin{tabular}{@{}llr*{12}{r}@{}}
\toprule
 & & & \multicolumn{3}{c}{\textsc{SIGReg-OOB} (Ours)} & \multicolumn{3}{c}{\sigreg{}} & \multicolumn{3}{c}{\contooob{} (Ours)} & \multicolumn{3}{c}{SimCLR} \\
\cmidrule(lr){4-6}\cmidrule(lr){7-9}\cmidrule(lr){10-12}\cmidrule(l){13-15}
Dataset & Encoder & Original & $B=2$ & $B=8$ & $B=64$ & $B=2$ & $B=8$ & $B=64$ & $B=2$ & $B=8$ & $B=64$ & $B=2$ & $B=8$ & $B=64$ \\
\midrule
BloodMNIST & S (22M) & 92.0 & \textbf{+6.6} & \textbf{+6.7} & \textbf{+6.8} & -6.5 & \textbf{+6.7} & +6.0 & \textbf{+6.1} & +6.2 & +5.8 & +1.6 & \textbf{+6.7} & \textbf{+6.5} \\
 & B (86M) & 90.1 & \textbf{+8.8} & +8.6 & \textbf{+8.6} & -5.7 & \textbf{+8.7} & +8.4 & \textbf{+8.0} & +8.4 & +8.1 & +4.1 & \textbf{+8.7} & \textbf{+8.7} \\
 & L (304M) & 89.4 & \textbf{+9.3} & \textbf{+9.4} & \textbf{+9.3} & -4.6 & \textbf{+9.4} & +9.0 & \textbf{+9.1} & +8.9 & \textbf{+9.5} & +2.5 & \textbf{+9.4} & +9.2 \\
 & G (1.1B) & 93.0 & \textbf{+5.3} & +5.1 & \textbf{+5.8} & -6.8 & \textbf{+5.8} & +5.3 & +2.8 & -5.6 & +3.7 & \textbf{+3.6} & \textbf{+5.5} & \textbf{+5.7} \\
\addlinespace[2pt]
OrganAMNIST & S (22M) & 85.7 & \textbf{+7.1} & +6.1 & \textbf{+5.9} & +0.9 & \textbf{+8.0} & +4.1 & \textbf{+8.7} & \textbf{+8.2} & \textbf{+8.1} & -15.9 & +4.6 & +4.7 \\
 & B (86M) & 84.0 & \textbf{+10.4} & +8.6 & \textbf{+7.1} & -0.4 & \textbf{+8.7} & +4.6 & \textbf{+10.6} & \textbf{+10.5} & \textbf{+10.3} & -18.0 & +7.1 & +6.3 \\
 & L (304M) & 79.3 & \textbf{+15.1} & \textbf{+14.3} & \textbf{+12.0} & +0.8 & +13.0 & +9.5 & \textbf{+14.0} & \textbf{+15.2} & \textbf{+15.6} & -15.0 & +10.9 & +10.5 \\
 & G (1.1B) & 80.0 & \textbf{+14.0} & \textbf{+13.1} & \textbf{+11.6} & +10.1 & +12.3 & +9.6 & \textbf{+13.5} & +11.2 & +11.9 & -8.9 & \textbf{+11.3} & \textbf{+12.1} \\
\addlinespace[2pt]
OrganCMNIST & S (22M) & 81.2 & \textbf{+1.7} & \textbf{+4.5} & \textbf{+4.0} & -5.9 & +4.1 & +2.7 & \textbf{+6.8} & \textbf{+6.8} & \textbf{+4.7} & -12.2 & +3.2 & +0.9 \\
 & B (86M) & 77.3 & \textbf{+6.5} & \textbf{+7.7} & \textbf{+6.6} & -0.1 & +7.5 & +6.0 & \textbf{+9.3} & \textbf{+11.9} & \textbf{+9.1} & -16.5 & +7.3 & +5.1 \\
 & L (304M) & 76.3 & \textbf{+8.3} & \textbf{+7.9} & \textbf{+8.3} & -0.2 & +7.7 & +5.6 & \textbf{+10.1} & \textbf{+13.9} & \textbf{+12.3} & -7.0 & +7.2 & +4.9 \\
 & G (1.1B) & 74.1 & \textbf{+11.1} & \textbf{+10.6} & \textbf{+8.8} & +10.7 & \textbf{+10.6} & +7.7 & \textbf{+13.4} & \textbf{+12.0} & \textbf{+12.3} & -3.2 & +11.3 & +6.6 \\
\addlinespace[2pt]
PathMNIST & S (22M) & 94.8 & \textbf{+2.5} & \textbf{+2.6} & \textbf{+2.9} & -7.6 & +1.0 & +1.9 & \textbf{+1.3} & \textbf{+2.2} & \textbf{+2.1} & -9.4 & +1.8 & +1.9 \\
 & B (86M) & 94.0 & \textbf{+3.5} & \textbf{+3.6} & \textbf{+3.6} & -9.6 & +2.5 & +2.6 & \textbf{+2.2} & +2.9 & +2.7 & -6.0 & \textbf{+3.2} & \textbf{+3.6} \\
 & L (304M) & 93.7 & \textbf{+4.0} & \textbf{+3.2} & \textbf{+4.6} & -14.1 & +3.0 & +3.3 & \textbf{+2.3} & +2.3 & +3.1 & -3.7 & \textbf{+3.8} & \textbf{+3.8} \\
 & G (1.1B) & 95.5 & \textbf{+1.5} & \textbf{+2.2} & \textbf{+2.2} & -13.9 & +0.9 & +1.2 & \textbf{-1.5} & -6.7 & -1.8 & -2.3 & \textbf{+1.6} & \textbf{+2.3} \\
\bottomrule
\end{tabular}%
}

\end{table}

\Cref{tab:continued} shows that the OOB methods outperform their
corresponding in-batch variants across most configurations. The advantage is clearest at small batches, where OOB
regularization improves on the pretrained initialization while in-batch
training can degrade representation quality. These results support using OOB regularization to
adapt large pretrained encoders when memory limits the training batch size.
Complete scores, linear-probe results, and all batch sizes are in \cref{app:continued}.

\subsection{Performance under a memory budget}
\label{sec:memory-results}

\begin{figure}[h]
  \centering
  \includegraphics[width=\textwidth]{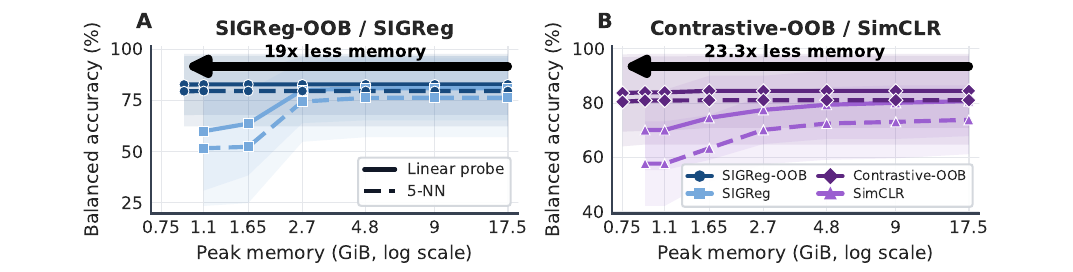}
  \caption{Best classification performance under GPU-memory budgets.}
  \label{fig:memory}
\end{figure}

At the same batch size, OOB methods use the same GPU memory
as their in-batch counterparts. Their advantage is that they perform well
with much smaller batches, reducing the memory needed to obtain useful
representations. We compare the best classification performance achievable
within each GPU-memory budget on the datasets in \cref{sec:datasets}. \Cref{fig:memory} shows that OOB methods can match or exceed the performance
of their in-batch counterparts while using substantially less GPU memory
within the evaluated range, making strong representations
accessible under tighter memory budgets. Training times and detailed resource
measurements are provided in \cref{app:compute}.

\section{Conclusion and Future Work}
\label{sec:conclusion}
We introduced out-of-batch (OOB) regularization, which decouples population-level representation regularization from minibatch optimization through periodically constructed dataset-level targets. Across contrastive and non-contrastive objectives, OOB more faithfully recovers the full-dataset regularization signal and improves representation quality, particularly in cases where small minibatches poorly approximate the full population. Our results show that the active minibatch does not need to define the population over which representation geometry is regularized.

OOB replaces minibatch sampling error with a different trade-off: dataset-level state must be stored and periodically refreshed as the encoder evolves. Understanding this trade-off at larger scales remains an important direction. More broadly, our results motivate separating globally coupled representation objectives from the minibatch updates used to train the encoder, with regularizer-specific constructions providing a possible route to population-aware learning beyond the objectives studied here.

\subsection*{AI use statement}

We used generative AI tools to polish the manuscript, assist in searching for
relevant literature, and help write code. We did not use generative AI to
generate research ideas or propose experiments. We reviewed the AI-assisted
text and references, and independently inspected and executed the AI-assisted
code to verify its correctness. We take responsibility for the final content
of this work, including all text, claims, and artifacts produced with the
assistance of generative AI.

\subsection*{Reproducibility statement}

We describe the proposed algorithms in the main text and provide detailed
architectures, training configurations, augmentations, hyperparameters, and evaluation protocols in the appendix to support independent
reproduction of our results. We also document the hyperparameter-selection
procedure and computational costs. We plan to release the code to further
facilitate reproduction and comparison with our methods.

\bibliography{references}
\bibliographystyle{arxiv_preprint}

\clearpage
\appendix
\section{Reproducibility details}
\label{app:reproducibility}

This section reports the data, optimization, augmentation, and evaluation
settings used in the main experiments. Classification and retrieval results
are aggregated over independent training seeds. Within each matched
OOB/in-batch classification pair, we
preserve the data split, augmentations, number of views, encoder, projector,
optimizer, schedule, and downstream evaluator.

\subsection{Datasets and evaluation protocols}
\label{app:datasets}

\paragraph{Classification.}
The datasets used in each collection are summarized in
\cref{tab:supp-datasets}.

\begin{table}[htbp]
  \centering
  \caption{Datasets used in the classification and oracle experiments.}
  \label{tab:supp-datasets}
  \footnotesize
  \setlength{\tabcolsep}{4pt}
  \begin{tabularx}{\textwidth}{@{}p{0.22\textwidth}X@{}}
    \toprule
    Collection & Domains \\
    \midrule
    Natural images & CIFAR-10, CIFAR-100, DTD, ImageNet-100, Imagenette,
    Imagewoof, STL-10 \\
    2D Biomedical & BloodMNIST, BreastMNIST, DermaMNIST, OCTMNIST, OrganAMNIST, OrganCMNIST, OrganSMNIST, PathMNIST,
    PneumoniaMNIST, TissueMNIST, BBBC010, plankton(private) \\
    3D Biomedical & AdrenalMNIST, FractureMNIST, NoduleMNIST, OrganMNIST, SynapseMNIST, VesselMNIST \\
    \bottomrule
  \end{tabularx}
\end{table}

For frozen evaluation, we extract deterministic, $\ell_2$-normalized backbone
features and use a fixed, shuffled 70/30 train/evaluation split using one seed.
We report balanced accuracy from cosine $k$-NN ($k\in\{1,5,20\}$) and from a multinomial logistic-regression probe (linear probe). The latter is fit
after standardizing each feature coordinate, with a maximum of 1000 iterations.
Labels are never used during self-supervised pretraining.

\paragraph{Instance retrieval.}
The retrieval study uses COIL-100 (7200 images of 100 objects), UKBench
(10,200 images of 2550 objects), and INRIA Holidays (1491 images grouped into
500 instances). Retrieval uses cosine similarity between normalized frozen
features, with the query itself removed from the gallery. For a query $q$ with
relevant gallery set $G_q$, average precision is
\begin{equation}
  \operatorname{AP}(q)=\frac{1}{|G_q|}
  \sum_{k=1}^{|\mathcal G|}P_q(k)\,
  \mathbf{1}\!\left[\pi_q(k)\in G_q\right],
\end{equation}
where $P_q(k)$ is precision at rank $k$ and $\pi_q(k)$ is the item at that
rank. mAP averages AP over queries; Recall@$K$ is the fraction of queries with
at least one relevant item among the first $K$ results. We run two different protocols for our experiments. The transductive
protocol pretrains and retrieves on the same unlabeled collection. The held-out
protocol separates instance identities with a fixed 80/20 split using one seed. All retrieval methods use the same 2D encoder, image transformations,
optimizer, and evaluator as the classification experiments, with four
augmented views per instance for every method.

\subsection{Architectures, optimization, and augmentation}
\label{app:training-configuration}

The 2D experiments use a ViT-Small/16 encoder with width 384,
12 transformer blocks, and six attention heads. Inputs are $224\times224$.
The 3D experiments use a 3D ResNet-18 and $64^3$ inputs. All methods
train for 100 epochs with AdamW, learning rate $10^{-4}$, weight decay 0.05,
gradient clipping at norm 1, and a per-step cosine schedule ending at
$10^{-6}$. We use the PyTorch AdamW defaults
$(\beta_1,\beta_2)=(0.9,0.999)$ and $\epsilon=10^{-8}$. From-scratch training
uses fp32, with each experiment run on a single NVIDIA L40S GPU.

\begin{table}[htbp]
  \centering
  \caption{Augmentation pipelines used for self-supervised training.}
  \label{tab:supp-augmentations}
  \footnotesize
  \setlength{\tabcolsep}{3.5pt}
  \renewcommand{\arraystretch}{1.16}
  \begin{tabularx}{\textwidth}{@{}l
    >{\raggedright\arraybackslash}X
    >{\raggedright\arraybackslash}X@{}}
    \toprule
    Modality & Spatial transforms & Appearance and normalization \\
    \midrule
    2D & Random resized crop to $224$ (scale $0.2$--$1.0$); horizontal flip
       & Color jitter $(0.4,0.4,0.4,0.1)$, $p=0.8$; grayscale, $p=0.2$;
         Gaussian blur (kernel 23, $\sigma\in[0.1,2.0]$); ImageNet normalization \\
    \addlinespace[2pt]
    3D & Random resized crop to $64^3$ (scale $0.5$--$1.0$); independent axis
         flips, $p=0.5$; random $90^\circ$ rotation in one plane, $p=0.5$
       & Intensity shift in $[-0.1,0.1]$; contrast in $[0.8,1.2]$; Gaussian
         noise ($\sigma=0.1$), $p=0.3$; $3^3$ average-pool blur, $p=0.3$ \\
    \bottomrule
  \end{tabularx}
\end{table}

\subsection{Method-specific hyperparameters and cost}
\label{app:method-hyperparameters}

\begin{table}[htbp]
  \centering
  \caption{Method-specific settings used in the principal experiments.}
  \label{tab:supp-hyperparameters}
  \footnotesize
  \setlength{\tabcolsep}{4pt}
  \begin{tabularx}{\textwidth}{@{}lX@{}}
    \toprule
    Method & Configuration \\
    \midrule
    SIGReg-OOB & Three-layer MLP projector with
    2048-dimensional hidden and output layers, LayerNorm and ReLU after each
    non-final layer;
    $M=256$ projections; $(\lambda_{\mathrm{align}},\lambda_{\mathrm{ac}})
    =(5,1)$; projection directions resampled at every refresh. \\
    \sigreg{} & The same 2048-dimensional projector \\
    \contooob{} & Two-layer ReLU projector with one backbone-width hidden layer
    and 128-dimensional output anchors;
    $(\lambda_{\mathrm{align}},\lambda_{\mathrm{ac}})=(1,1)$; 20 anchor
    updates with Adam at learning rate 0.05; sampled anchor subset 256;
    $\tau_u=2$, $\lambda_f=0.1$, and previous-cache mixture weight 0.5. \\
    SimCLR & The same two-layer 128-dimensional projector, normalized output,
    four augmented views in the classification experiments, and temperature
    0.5. \\
    MoCoV2 & Two-layer 128-dimensional query/key projectors, four augmented
    views, queue size 4096, encoder momentum 0.999, and temperature 0.2. \\
    \bottomrule
  \end{tabularx}
\end{table}

Both OOB methods use a fixed refresh period of 0.5 epoch. Refresh inference is
batched independently of online training, with batch size 64 in 2D and 16 in
3D. SIGReg-OOB requires one deterministic dataset pass followed by
$O(MN\log N)$ CPU sorting. Its persistent fp32 cache contains $NM$ scalars.
\contooob{} performs 20 stochastic anchor updates, each using all pairs within
a sampled subset of at most 256 anchors, while its persistent cache contains $128N$
scalars. The caches live in CPU memory, and only the rows indexed by the active
minibatch are transferred to the GPU. At $N=13{,}000$, these caches occupy
12.7 MiB and 6.3 MiB, respectively. Cache construction adds wall time but does
not increase the online training batch's activation memory. Details about training setup of baselines and our methods are reported in \Cref{tab:supp-hyperparameters}, while the augmentations we use are described in \Cref{tab:supp-augmentations}.

\subsection{Ablations of the SIGReg-OOB configuration}
\label{app:hparam-ablations}

We ablate the components of SIGReg-OOB on seven 2D natural/scientific domains
and batches 1--64. The study comprises 1,792 evaluations from 39 configurations
over 48 dataset--batch matched experiments, covering projection resampling, refresh period,
the alignment and anti-collapse weights, and the number of projections. Since
the sweep is not a full Cartesian product, we form controlled slices in which
one factor changes while the remaining settings are held fixed. Within each
slice, we retain only dataset--batch experiments shared by every variant.

The resampling, refresh-period, and alignment-weight ablations use the common
configuration $p=0.5$, $\lambda_{\mathrm{align}}=5$,
$\lambda_{\mathrm{ac}}=1$, and $M=256$, omitting the factor being varied. The
anti-collapse-weight slice fixes $(p,\lambda_{\mathrm{align}},M)=(2,25,512)$,
and the projection-count slice fixes
$(p,\lambda_{\mathrm{align}},\lambda_{\mathrm{ac}})=(2,25,5)$; projection
resampling is enabled in both.

\begin{figure}[htbp]
  \centering
  \includegraphics[width=0.96\textwidth]{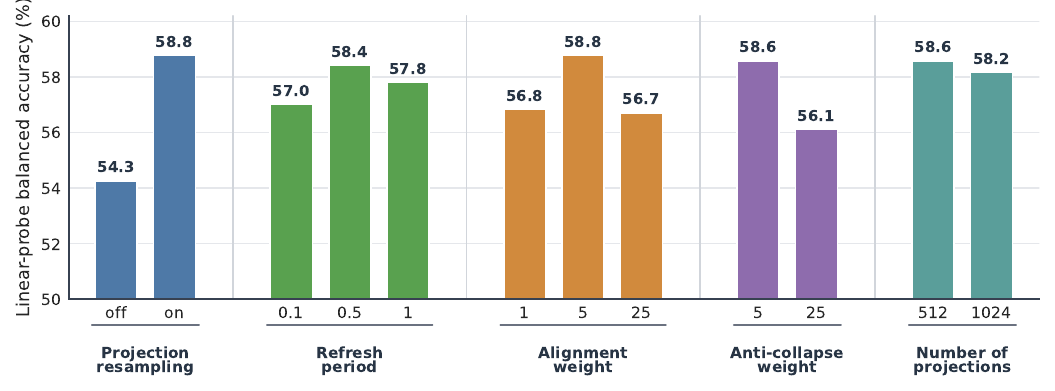}
  \caption{Controlled ablations of SIGReg-OOB. Each factor varies within a
  matched slice, and bars report mean linear-probe balanced accuracy over the
  dataset--batch cells shared by all values in that group.}
  \label{fig:supp-hparam-ablations}
\end{figure}

\Cref{fig:supp-hparam-ablations} shows mean downstream linear-probe balanced
accuracy. Projection resampling produces the largest improvement. A refresh
period of 0.5 epoch performs best, while one epoch remains close and overly
frequent refreshes are not beneficial. The intermediate alignment weight is
preferable to either extreme, and increasing the anti-collapse weight or the
number of projections does not improve the matched slices. These results
indicate that maintaining diverse, periodically updated targets and balancing
the two loss terms matter more than increasing the raw number of projections.

\section{Oracle-gradient protocol}
\label{app:oracle}

The oracle study differentiates the projected-Gaussian anti-collapse loss with
respect to each probe representation. Consequently, every ``gradient'' is a
vector in representation space, not the scalar loss and not the full network
parameter gradient. The full-dataset reference recomputes ranks and Gaussian
quantile assignments from the current representations of every training
sample. The minibatch estimator computes the same construction from a sampled
batch, whereas the cached estimator reuses full-dataset assignments made at an
earlier encoder state. At every cache age, the reference itself is recomputed
at the current encoder state; cached agreement is therefore measured against
the current population objective rather than against the reference at refresh.

The canonical experiment uses all 2D and 3D datasets considered in this study.
We evaluate checkpoints at epochs 0, 50, and 100 over three seeds. Each condition
uses 16 probe samples, 16 sampled contexts, and 256 projection directions.
Minibatch sizes are 1--64 and cache ages are
$s\in\{0,1,2,4,8,16,32,64,128\}$ optimizer updates. Direction is measured by
gradient cosine similarity, and scale-sensitive disagreement by relative
root-mean-square error.

At $B=1$, mean cosine agreement at epochs 0, 50, and 100 is respectively
$-0.047$, $-0.417$, and $-0.362$; at $B=64$ it rises to $0.972$, $0.903$, and
$0.939$. A fresh cache agrees exactly by construction. At age 128, mean cached
agreement at the default learning rate is $0.453$, $0.720$, and $0.912$ across
the three checkpoints. At the matched low-rate control ($10^{-6}$), the values
are $0.692$, $0.909$, and $0.906$, with the largest improvements at epochs 0
and 50. Cache lifetime is therefore controlled by how rapidly the
encoder is updated, not by update count alone.

\begin{figure}[htbp]
  \centering
  \includegraphics[width=0.96\textwidth]{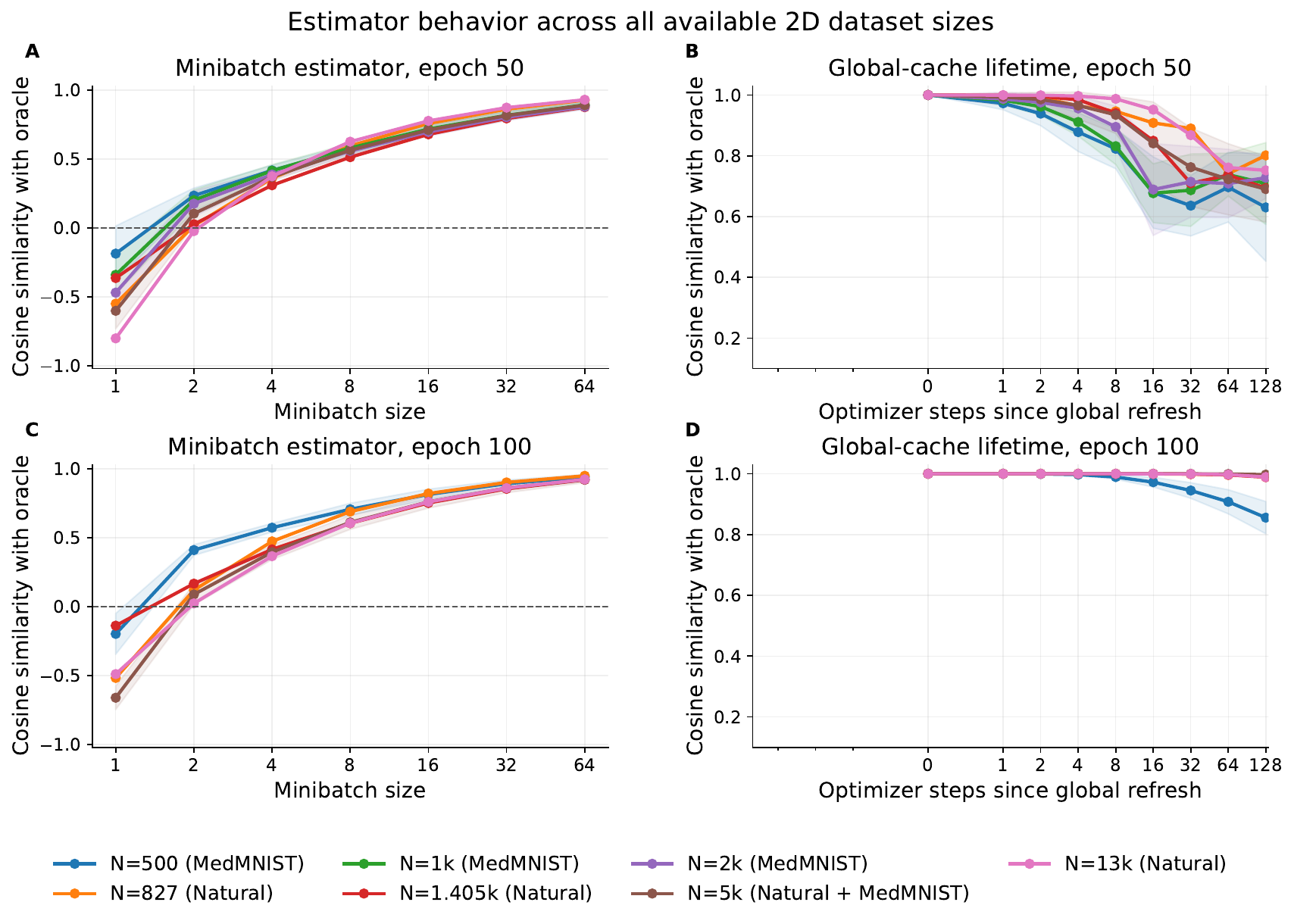}
  \caption{Robustness of the oracle comparison to dataset size. The MedMNIST
  sweep uses 500, 1k, 2k, and 5k training examples. Panels A and C evaluate the
  minibatch estimator, while B and D follow global-cache lifetime at epochs 50
  and 100, respectively.}
  \label{fig:supp-oracle-size}
\end{figure}

\Cref{fig:supp-oracle-size} verifies that the ordering between limited batch
context and fresh dataset-level context is not an artifact of a single dataset
size. The oracle remains a reference for estimating this chosen
finite-population regularizer.

\section{Detailed frozen-classification results}
\label{app:classification}

\begin{figure}[htbp]
  \centering
  \includegraphics[width=0.96\textwidth]{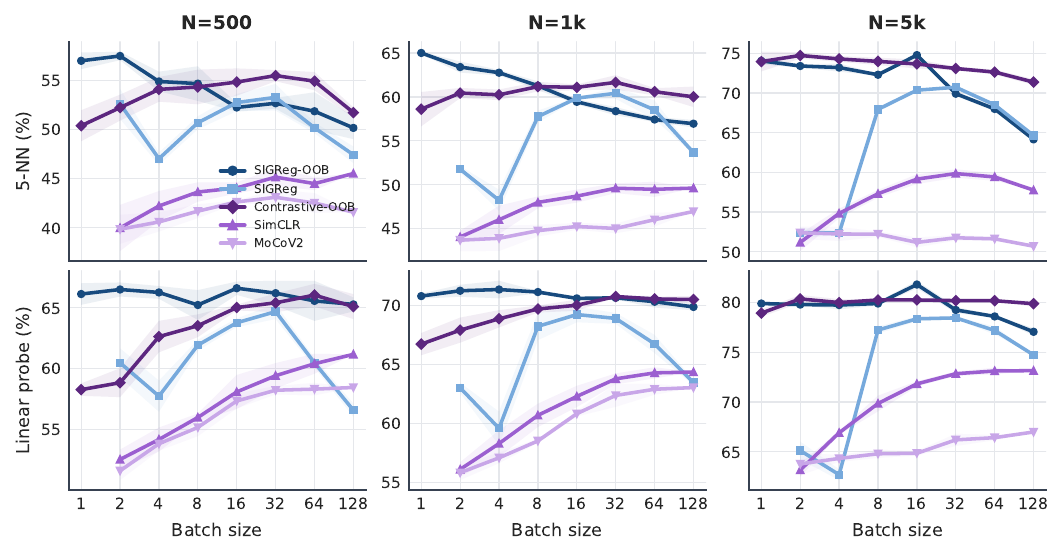}
  \caption{Frozen classification across MedMNIST training-set sizes. Shading
  denotes one standard deviation across training seeds.}
  \label{fig:supp-size-sweep}
\end{figure}

\Cref{fig:supp-size-sweep} summarizes the controlled 2D MedMNIST size sweep.
The advantage of OOB regularization is largest at small batches and persists
as the number of training examples grows. Tables
\ref{tab:supp-medmnist-datasets} and \ref{tab:supp-classification-datasets}
report the per-dataset results after averaging over a common set of batch
sizes. The 2D biomedical and 3D tables average
$B\in\{2,4,8,16,32,64\}$, while the natural-image table averages
$B\in\{2,4,8,16,32\}$. The full ImageNet-100 row instead averages its
available batches 64 and 128. Each entry gives 5-NN and linear-probe balanced
accuracy, in that order.

Across the controlled size sweep, the OOB variants retain their strongest
advantages in the low-batch regime rather than relying on a particular number
of training examples. The per-dataset results below also show that these gains
are distributed across domains rather than being driven by a single task.

\begin{table}[p]
  \centering
  \caption{Per-dataset 2D MedMNIST balanced accuracy (\%) across training-set
  sizes, averaged over batch sizes. Entries are linear probe results. Within
  each method family, the best result is bold; the best result overall is
  underlined.}
  \label{tab:supp-medmnist-datasets}
  \scriptsize
  \setlength{\tabcolsep}{2pt}
  \renewcommand{\arraystretch}{0.98}
  \textbf{(a) $N=500$}\par\vspace{2pt}
  \begin{tabularx}{\linewidth}{@{}l*{2}{>{\centering\arraybackslash}X}|*{3}{>{\centering\arraybackslash}X}|*{6}{>{\centering\arraybackslash}X}@{}}
\toprule
Dataset & SIGReg & \shortstack{SIGReg-\\[-1pt]OOB} & SimCLR & MoCoV2 & \shortstack{\fontsize{6}{6.4}\selectfont Contrastive\\[-1pt]\fontsize{6}{6.4}\selectfont -OOB} & VICReg & IConE & \shortstack{IConE-\\[-1pt]SIGReg} & BYOL & SwAV & DINO \\
\midrule
BloodMNIST & 82.4 & \underline{\textbf{84.4}} & 68.9 & 68.1 & \textbf{78.2} & 76.3 & 81.1 & 80.8 & 72.9 & 75.7 & 72.6 \\
BreastMNIST & 71.3 & \underline{\textbf{72.5}} & 66.6 & 65.2 & \textbf{72.0} & 70.9 & 68.8 & 68.5 & 67.8 & 68.1 & 65.5 \\
DermaMNIST & 54.7 & \underline{\textbf{60.0}} & 50.9 & 48.6 & \textbf{59.5} & 52.4 & 45.2 & 45.2 & 49.1 & 49.3 & 46.3 \\
OCTMNIST & 48.2 & \underline{\textbf{52.6}} & 41.2 & 39.3 & \textbf{41.8} & 48.9 & 42.9 & 44.2 & 41.5 & 43.8 & 38.2 \\
OrganAMNIST & 61.0 & \textbf{70.7} & 54.3 & 52.8 & \textbf{66.9} & 61.0 & \underline{73.1} & 72.6 & 55.2 & 56.8 & 52.5 \\
OrganCMNIST & 58.9 & \textbf{65.6} & 51.2 & 47.8 & \textbf{63.5} & 64.3 & 70.9 & \underline{71.5} & 50.2 & 51.6 & 53.0 \\
OrganSMNIST & 55.3 & \underline{\textbf{56.4}} & 46.8 & 44.4 & \textbf{56.0} & 52.9 & 55.2 & 55.8 & 47.2 & 48.0 & 46.6 \\
PathMNIST & 70.4 & \underline{\textbf{76.5}} & 68.0 & 70.9 & \textbf{74.4} & 74.0 & 66.0 & 65.5 & 68.2 & 72.1 & 64.2 \\
PneumoniaMNIST & 86.7 & \underline{\textbf{88.7}} & 85.4 & 84.2 & \textbf{88.3} & 88.0 & 85.1 & 85.4 & 87.1 & 86.2 & 82.6 \\
TissueMNIST & 28.4 & \underline{\textbf{34.1}} & 27.1 & \textbf{31.1} & 30.3 & 30.0 & 27.9 & 28.7 & 29.9 & 30.9 & 28.5 \\
\bottomrule
\end{tabularx}

  \vspace{4pt}
  \textbf{(b) $N=1$k}\par\vspace{2pt}
  \begin{tabularx}{\linewidth}{@{}l*{2}{>{\centering\arraybackslash}X}|*{3}{>{\centering\arraybackslash}X}|*{6}{>{\centering\arraybackslash}X}@{}}
\toprule
Dataset & SIGReg & \shortstack{SIGReg-\\[-1pt]OOB} & SimCLR & MoCoV2 & \shortstack{\fontsize{6}{6.4}\selectfont Contrastive\\[-1pt]\fontsize{6}{6.4}\selectfont -OOB} & VICReg & IConE & \shortstack{IConE-\\[-1pt]SIGReg} & BYOL & SwAV & DINO \\
\midrule
BloodMNIST & 86.5 & \underline{\textbf{91.7}} & 79.6 & 79.9 & \textbf{90.6} & 84.1 & 84.2 & 84.9 & 81.7 & 82.0 & 77.6 \\
BreastMNIST & 72.4 & \underline{\textbf{74.6}} & 69.9 & 67.5 & \textbf{71.5} & 71.2 & 72.7 & 71.3 & 71.0 & 70.6 & 66.5 \\
DermaMNIST & 52.2 & \underline{\textbf{58.0}} & 50.2 & 48.4 & \textbf{57.7} & 51.9 & 48.3 & 48.4 & 48.8 & 48.1 & 46.3 \\
OCTMNIST & 53.6 & \underline{\textbf{59.5}} & 45.4 & 45.4 & \textbf{51.7} & 53.4 & 45.5 & 44.8 & 44.6 & 48.9 & 41.5 \\
OrganAMNIST & 72.1 & \textbf{77.2} & 61.4 & 61.1 & \textbf{76.4} & 67.4 & \underline{77.4} & 76.9 & 59.6 & 61.7 & 53.0 \\
OrganCMNIST & 70.4 & \textbf{76.5} & 60.4 & 57.3 & \textbf{76.8} & 71.3 & \underline{80.1} & 79.2 & 59.4 & 61.1 & 57.0 \\
OrganSMNIST & 59.5 & \textbf{60.8} & 49.2 & 48.9 & \underline{\textbf{61.8}} & 54.5 & 59.4 & 59.5 & 48.8 & 49.2 & 47.4 \\
PathMNIST & 67.7 & \underline{\textbf{79.0}} & 67.6 & 67.1 & \textbf{76.9} & 69.1 & 56.2 & 55.2 & 66.2 & 64.5 & 60.1 \\
PneumoniaMNIST & 88.0 & \underline{\textbf{91.4}} & 84.1 & 81.3 & \textbf{89.5} & 85.4 & 86.8 & 85.7 & 84.9 & 84.4 & 80.8 \\
TissueMNIST & 35.6 & \textbf{41.3} & 34.4 & 32.4 & \underline{\textbf{41.7}} & 34.9 & 31.2 & 33.3 & 33.4 & 32.8 & 30.8 \\
\bottomrule
\end{tabularx}

  \vspace{4pt}
  \textbf{(c) $N=2$k}\par\vspace{2pt}
  \begin{tabularx}{\linewidth}{@{}l*{2}{>{\centering\arraybackslash}X}|*{3}{>{\centering\arraybackslash}X}|*{6}{>{\centering\arraybackslash}X}@{}}
\toprule
Dataset & SIGReg & \shortstack{SIGReg-\\[-1pt]OOB} & SimCLR & MoCoV2 & \shortstack{\fontsize{6}{6.4}\selectfont Contrastive\\[-1pt]\fontsize{6}{6.4}\selectfont -OOB} & VICReg & IConE & \shortstack{IConE-\\[-1pt]SIGReg} & BYOL & SwAV & DINO \\
\midrule
BloodMNIST & 90.1 & \underline{\textbf{95.4}} & \textbf{82.1} & 80.2 & 81.9 & 85.9 & 88.3 & 88.0 & 83.7 & 83.4 & 77.2 \\
BreastMNIST & 73.6 & \underline{\textbf{75.6}} & 70.8 & 68.2 & \textbf{73.6} & 73.4 & 73.2 & 72.1 & 71.0 & 72.2 & 65.1 \\
DermaMNIST & 53.5 & \underline{\textbf{59.9}} & 52.5 & 48.4 & \textbf{52.8} & 53.8 & 49.0 & 48.3 & 47.8 & 45.2 & 43.3 \\
OCTMNIST & 58.1 & \underline{\textbf{66.0}} & \textbf{46.4} & 44.6 & 46.1 & 56.1 & 48.0 & 48.4 & 46.5 & 48.1 & 41.4 \\
OrganAMNIST & 75.0 & \underline{\textbf{81.4}} & \textbf{66.8} & 61.8 & 66.3 & 69.9 & 80.3 & 79.6 & 61.1 & 63.2 & 57.2 \\
OrganCMNIST & 76.2 & \underline{\textbf{82.6}} & \textbf{69.2} & 61.5 & 65.8 & 73.9 & 80.3 & 80.3 & 64.7 & 62.6 & 58.9 \\
OrganSMNIST & 62.6 & \underline{\textbf{69.3}} & 58.2 & 53.5 & \textbf{60.6} & 60.6 & 62.9 & 62.2 & 55.2 & 53.7 & 49.7 \\
PathMNIST & 75.5 & \underline{\textbf{85.3}} & \textbf{73.2} & 69.1 & 72.2 & 73.1 & 67.4 & 67.8 & 68.3 & 68.1 & 60.1 \\
PneumoniaMNIST & 91.7 & \underline{\textbf{94.2}} & \textbf{89.3} & 87.3 & 86.8 & 90.1 & 90.6 & 89.5 & 89.1 & 88.0 & 84.1 \\
TissueMNIST & 37.1 & \underline{\textbf{46.5}} & \textbf{34.2} & 32.5 & 33.1 & 38.0 & 30.1 & 30.4 & 32.2 & 31.8 & 27.4 \\
\bottomrule
\end{tabularx}

  \vspace{4pt}
  \textbf{(d) $N=5$k}\par\vspace{2pt}
  \begin{tabularx}{\linewidth}{@{}l*{2}{>{\centering\arraybackslash}X}|*{3}{>{\centering\arraybackslash}X}|*{6}{>{\centering\arraybackslash}X}@{}}
\toprule
Dataset & SIGReg & \shortstack{SIGReg-\\[-1pt]OOB} & SimCLR & MoCoV2 & \shortstack{\fontsize{6}{6.4}\selectfont Contrastive\\[-1pt]\fontsize{6}{6.4}\selectfont -OOB} & VICReg & IConE & \shortstack{IConE-\\[-1pt]SIGReg} & BYOL & SwAV & DINO \\
\midrule
BloodMNIST & 93.2 & \textbf{97.5} & 90.3 & 87.7 & \underline{\textbf{97.7}} & 88.4 & 91.3 & 92.3 & 87.7 & 90.0 & 80.9 \\
BreastMNIST & 73.3 & \underline{\textbf{75.4}} & 71.0 & 67.7 & \textbf{73.8} & 72.8 & 73.1 & 73.8 & 72.6 & 71.5 & 66.1 \\
DermaMNIST & 52.9 & \textbf{61.5} & 50.5 & 45.9 & \underline{\textbf{62.3}} & 50.6 & 47.6 & 46.1 & 45.5 & 44.3 & 40.4 \\
OCTMNIST & 63.5 & \textbf{76.5} & 56.7 & 51.5 & \underline{\textbf{77.1}} & 60.7 & 51.5 & 51.7 & 55.2 & 52.1 & 43.2 \\
OrganAMNIST & 81.8 & \textbf{88.2} & 75.7 & 69.8 & \underline{\textbf{89.1}} & 76.1 & 80.7 & 82.3 & 67.0 & 64.4 & 56.6 \\
OrganCMNIST & 78.9 & \textbf{85.0} & 76.5 & 68.4 & \underline{\textbf{86.1}} & 74.9 & 79.8 & 80.7 & 69.6 & 65.2 & 59.5 \\
OrganSMNIST & 68.4 & \textbf{75.0} & 66.0 & 61.3 & \underline{\textbf{76.6}} & 64.4 & 64.6 & 68.3 & 58.1 & 58.5 & 51.0 \\
PathMNIST & 79.8 & \textbf{92.2} & 76.7 & 73.8 & \underline{\textbf{92.9}} & 74.6 & 73.3 & 72.5 & 70.9 & 72.3 & 62.5 \\
PneumoniaMNIST & 92.4 & \textbf{95.8} & 89.5 & 87.2 & \underline{\textbf{96.0}} & 89.1 & 89.2 & 87.6 & 89.5 & 87.3 & 82.0 \\
TissueMNIST & 39.6 & \textbf{49.7} & 36.3 & 34.8 & \underline{\textbf{50.5}} & 38.0 & 32.8 & 32.3 & 33.9 & 32.6 & 28.6 \\
\bottomrule
\end{tabularx}

\end{table}

The natural-image and 3D results in
\cref{tab:supp-classification-datasets} broaden the comparison beyond the
controlled 2D size sweep while retaining the same frozen-evaluation protocol.

\begin{table}[!ht]
  \centering
  \caption{Per-dataset balanced accuracy (\%) on natural/scientific 2D and
  3D MedMNIST datasets, averaged over batch sizes. Entries are 5-NN / linear
  probe. Within each method family, the best result is bold; the best result
  overall is underlined.}
  \label{tab:supp-classification-datasets}
  \footnotesize
  \setlength{\tabcolsep}{2pt}
  \renewcommand{\arraystretch}{1.02}
  \textbf{(a) Natural/scientific 2D}\par\vspace{2pt}
  \begin{tabularx}{\textwidth}{@{}l*{2}{>{\centering\arraybackslash}X}|*{3}{>{\centering\arraybackslash}X}@{}}
\toprule
Dataset & SIGReg & SIGReg-OOB & SimCLR & MoCoV2 & \shortstack{Contrastive\\[-1pt]-OOB} \\
\midrule
BBBC010 & \underline{\textbf{90.9}}\,/\,\underline{\textbf{93.8}} & 90.4\,/\,91.5 & 89.9\,/\,90.9 & \textbf{90.8}\,/\,90.9 & 90.0\,/\,\textbf{91.2} \\
CIFAR-10 & 49.2\,/\,60.0 & \underline{\textbf{79.7}}\,/\,\underline{\textbf{82.2}} & \textbf{76.2}\,/\,\textbf{81.1} & 68.2\,/\,73.8 & 74.9\,/\,79.4 \\
CIFAR-100 & 20.7\,/\,33.5 & \underline{\textbf{42.2}}\,/\,\underline{\textbf{53.9}} & \textbf{39.1}\,/\,\textbf{53.1} & 34.6\,/\,46.9 & 33.4\,/\,49.4 \\
DTD & 17.5\,/\,25.2 & \textbf{26.0}\,/\,\textbf{32.7} & 26.5\,/\,35.8 & 22.3\,/\,29.1 & \underline{\textbf{32.4}}\,/\,\underline{\textbf{37.0}} \\
ImageNet-100 (full) & 44.3\,/\,53.9 & \underline{\textbf{52.4}}\,/\,\underline{\textbf{60.6}} & --\,/\,-- & --\,/\,-- & --\,/\,-- \\
ImageNet-100-5k & 39.3\,/\,54.4 & \textbf{55.9}\,/\,\textbf{67.8} & 53.8\,/\,65.3 & 49.2\,/\,63.2 & \underline{\textbf{63.9}}\,/\,\underline{\textbf{69.5}} \\
Imagenette & 39.5\,/\,51.9 & \textbf{58.2}\,/\,\textbf{66.6} & 56.5\,/\,65.9 & 45.8\,/\,58.4 & \underline{\textbf{66.0}}\,/\,\underline{\textbf{68.4}} \\
Imagewoof & 20.1\,/\,30.3 & \textbf{28.2}\,/\,\textbf{39.6} & 28.2\,/\,40.3 & 22.9\,/\,34.5 & \underline{\textbf{34.5}}\,/\,\underline{\textbf{42.1}} \\
Plankton & 41.3\,/\,42.8 & \textbf{42.6}\,/\,\underline{\textbf{45.8}} & \underline{\textbf{43.1}}\,/\,\textbf{45.6} & 40.5\,/\,42.6 & 42.9\,/\,45.5 \\
STL-10 & 43.4\,/\,52.4 & \textbf{61.5}\,/\,\textbf{67.5} & 59.9\,/\,66.4 & 50.3\,/\,59.6 & \underline{\textbf{64.6}}\,/\,\underline{\textbf{69.9}} \\
\bottomrule
\end{tabularx}

  \vspace{6pt}
  \textbf{(b) 3D MedMNIST}\par\vspace{2pt}
  \begin{tabularx}{\textwidth}{@{}l*{2}{>{\centering\arraybackslash}X}|*{3}{>{\centering\arraybackslash}X}@{}}
\toprule
Dataset & SIGReg & SIGReg-OOB & SimCLR & MoCoV2 & \shortstack{Contrastive\\[-1pt]-OOB} \\
\midrule
AdrenalMNIST3D & 54.4\,/\,57.6 & \textbf{57.1}\,/\,\textbf{64.7} & 55.6\,/\,60.0 & 56.5\,/\,62.9 & \underline{\textbf{58.8}}\,/\,\underline{\textbf{64.8}} \\
FractureMNIST3D & \textbf{37.7}\,/\,40.5 & 36.5\,/\,\textbf{41.3} & \underline{\textbf{37.8}}\,/\,\underline{\textbf{42.1}} & 37.6\,/\,41.4 & 36.6\,/\,39.2 \\
NoduleMNIST3D & 64.7\,/\,66.5 & \underline{\textbf{72.6}}\,/\,\underline{\textbf{76.7}} & 65.1\,/\,71.6 & 68.2\,/\,75.7 & \textbf{72.2}\,/\,\textbf{76.4} \\
OrganMNIST3D & 44.8\,/\,63.4 & \textbf{56.5}\,/\,\underline{\textbf{83.1}} & 42.2\,/\,60.4 & 55.8\,/\,\textbf{80.3} & \underline{\textbf{57.5}}\,/\,75.8 \\
SynapseMNIST3D & 55.5\,/\,56.8 & \underline{\textbf{57.6}}\,/\,\textbf{59.0} & \underline{\textbf{57.6}}\,/\,58.8 & 54.0\,/\,\underline{\textbf{60.4}} & 54.1\,/\,58.9 \\
VesselMNIST3D & 51.2\,/\,50.7 & \textbf{53.4}\,/\,\underline{\textbf{58.9}} & 52.3\,/\,52.2 & \underline{\textbf{54.7}}\,/\,\textbf{56.0} & 53.6\,/\,55.5 \\
\bottomrule
\end{tabularx}

\end{table}

\subsection{Representation-rank diagnostics}
\label{app:rank}

\begin{figure}[htbp]
  \centering
  \includegraphics[width=0.9\textwidth]{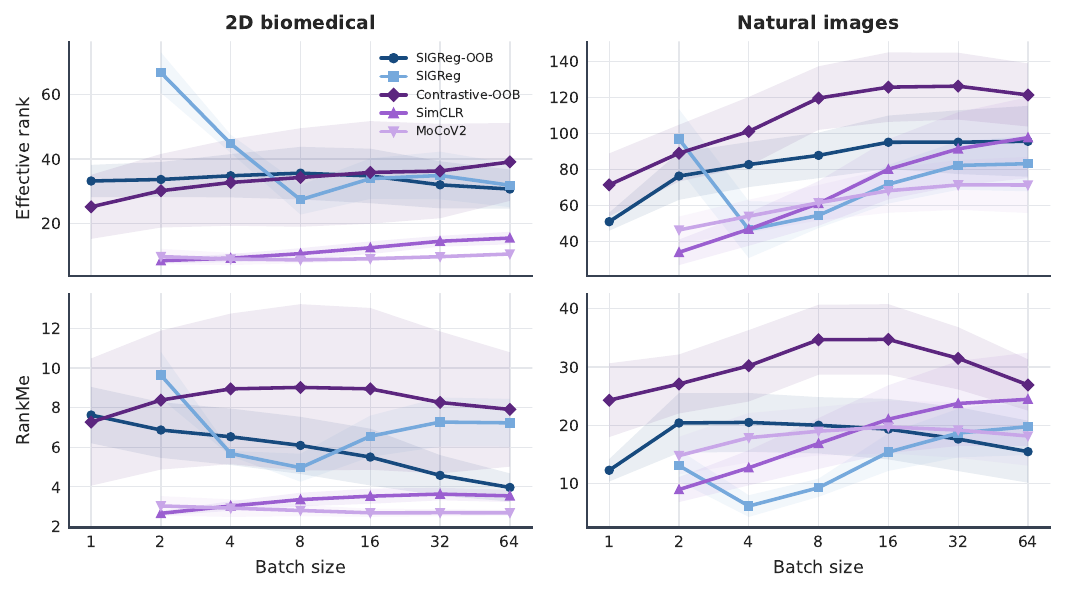}
  \caption{Effective rank and RankMe of frozen representations. Lines average
  over the common datasets and bands show standard errors across datasets.}
  \label{fig:supp-rank}
\end{figure}

\Cref{fig:supp-rank} provides a useful qualification. \contooob{} has much
higher rank than SimCLR and MoCoV2 in the smallest-batch regime, consistent
with stronger prevention of dimensional collapse. The Gaussian pair is more
mixed: SIGReg-OOB can have lower rank than \sigreg{} on 2D biomedical data
while matching or exceeding its downstream accuracy, and is comparable or
higher on natural images. Rank therefore diagnoses one aspect of collapse but
does not by itself explain semantic performance.

\section{Detailed class-imbalance results}
\label{app:imbalance}

\begin{figure}[htbp]
  \centering
  \includegraphics[width=0.96\textwidth]{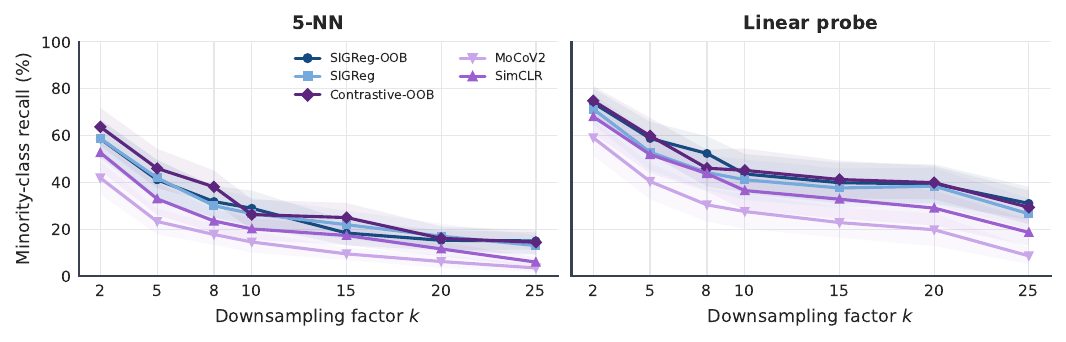}
  \caption{Minority-class recall for the controlled-imbalance 2D experiments
  at batch size 64: 5-NN (left) and linear probe (right).}
  \label{fig:supp-unbalanced-grid-2d}
\end{figure}

\begin{figure}[htbp]
  \centering
  \includegraphics[width=0.94\textwidth]{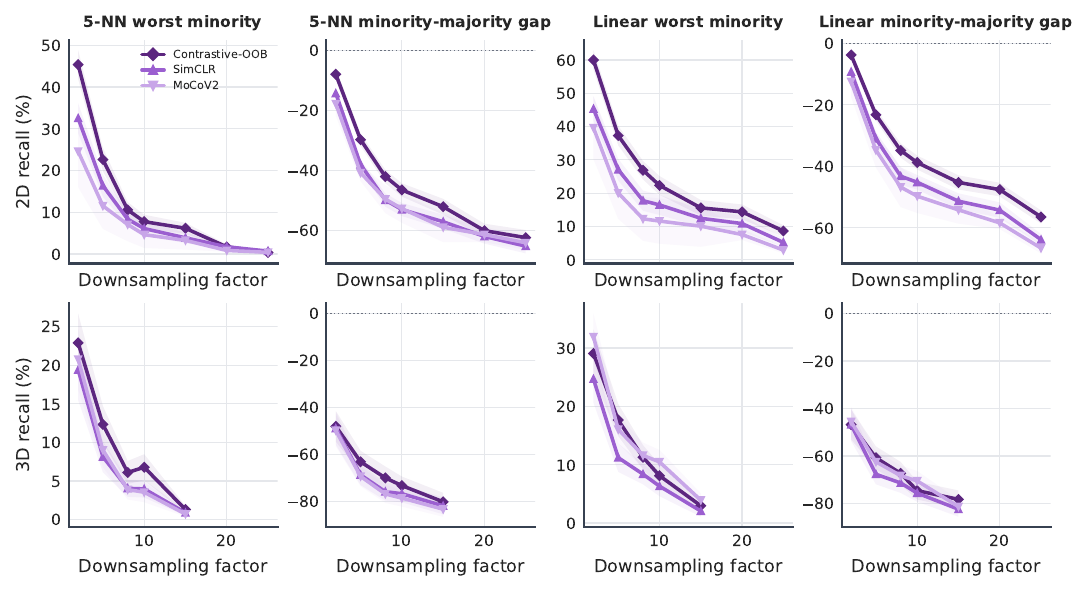}
  \caption{Worst-class recall and minority--majority recall gap for
  Contrastive-OOB, SimCLR, and MoCoV2. Lines average dataset--batch experiments and bands show standard errors across those results.}
  \label{fig:supp-unbalanced-worst}
\end{figure}

We generate controlled imbalance by retaining a $1/k$ fraction of the samples
from designated classes, for $k\in\{2,5,8,10,15,20,25\}$. The SSL algorithms
remain label-free. Class
labels are used only after pretraining to report macro recall over the
downsampled classes, worst-class recall, and the minority--majority recall
gap. \Cref{fig:supp-unbalanced-grid-2d} reports the 2D results at batch size 64,
while \cref{fig:supp-unbalanced-worst} condenses the more stringent tail
metrics for the contrastive methods.

The OOB gains are not restricted to average classification accuracy: they are
visible for the underrepresented classes and, in many settings, for the worst
class. Nevertheless, an empirical cache preserves the empirical population's
frequency weighting. The label-aware oracle diagnostic in the main paper
therefore separates two claims: OOB state can estimate a chosen population
objective more faithfully, but changing that objective to be class-balanced
requires an explicit weighting decision.

\section{Instance-retrieval results}
\label{app:retrieval}

\begin{figure}[h]
  \centering
  \includegraphics[width=\textwidth]{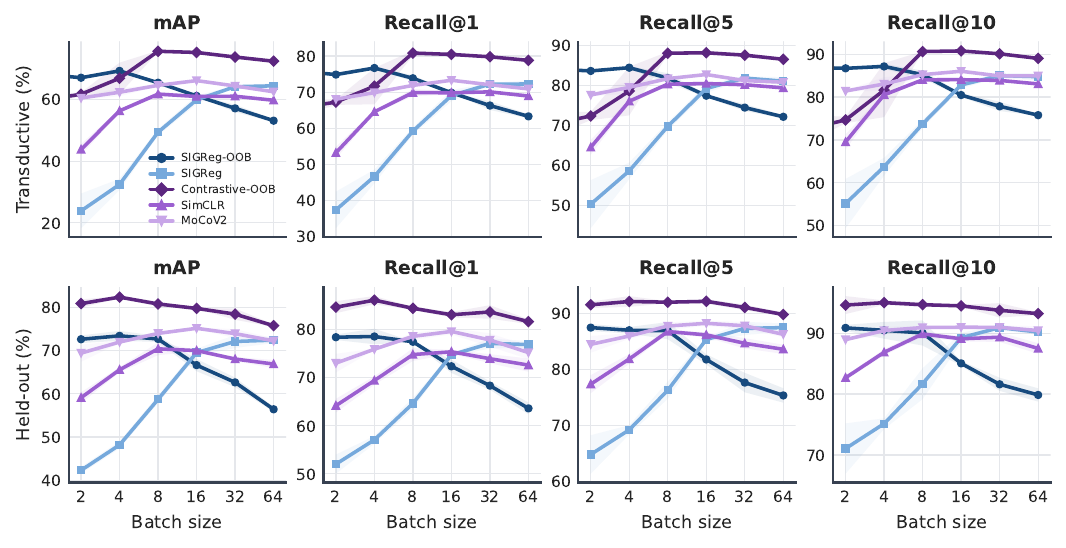}
  \caption{Transductive and held-out retrieval for mAP and Recall@$1/5/10$.
  For each seed, scores are first averaged over the three datasets; lines then
  show the three-seed mean and bands show one standard deviation.}
  \label{fig:supp-retrieval}
\end{figure}

\begin{table}[htbp]
  \centering
  \caption{Retrieval performance (\%), averaged over the three datasets,
  batches 2--64, and three training seeds within each protocol. Bold denotes an
  OOB result that exceeds its corresponding in-batch comparator: SIGReg-OOB
  versus SIGReg, or Contrastive-OOB versus both SimCLR and MoCoV2.}
  \label{tab:supp-retrieval}
  \scriptsize
  \resizebox{\textwidth}{!}{\begin{tabular}{llrrrrr}
\toprule
Protocol & Metric & SIGReg & SIGReg-OOB & SimCLR & Contrastive-OOB & MoCoV2 \\
\midrule
Transductive & mAP & 49.1 & \textbf{62.2} & 57.3 & \textbf{71.0} & 63.4 \\
 & R@1 & 59.5 & \textbf{70.8} & 66.1 & \textbf{76.5} & 71.0 \\
 & R@5 & 70.1 & \textbf{79.0} & 76.8 & \textbf{83.5} & 80.6 \\
 & R@10 & 74.2 & \textbf{82.2} & 80.8 & \textbf{86.1} & 84.3 \\
\midrule
Held-out & mAP & 60.6 & \textbf{67.4} & 66.7 & \textbf{79.7} & 72.8 \\
 & R@1 & 67.1 & \textbf{73.1} & 71.7 & \textbf{83.9} & 76.7 \\
 & R@5 & 78.4 & \textbf{82.7} & 83.4 & \textbf{91.4} & 86.7 \\
 & R@10 & 83.0 & \textbf{86.3} & 87.6 & \textbf{94.3} & 90.4 \\
\bottomrule
\end{tabular}
}
\end{table}

The held-out evaluation confirms that the transductive improvements
are not solely memorization of the training instances (\Cref{tab:supp-retrieval,fig:supp-retrieval}). \contooob{} gives the
highest average mAP and recall in both protocols, while SIGReg-OOB improves
over \sigreg{} on every aggregate retrieval metric.

\section{Continued pretraining with DINOv2}
\label{app:continued}

Continued-pretraining runs start from public DINOv2 ViT-S/B/L/G checkpoints
and use two augmented views for 100 epochs. The backbone and projector learning
rates are $10^{-5}$ and $10^{-4}$, respectively; weight decay is 0.05 and the
gradient norm is clipped at 1. OOB refreshes occur every 0.5 epoch. Refresh
batch sizes for ViT-S/B/L/G are 128, 64, 16, and 4 to accommodate their
different activation footprints. Experiments use an NVIDIA A100 GPU.

\begin{figure}[h]
  \centering
  \includegraphics[width=\textwidth]{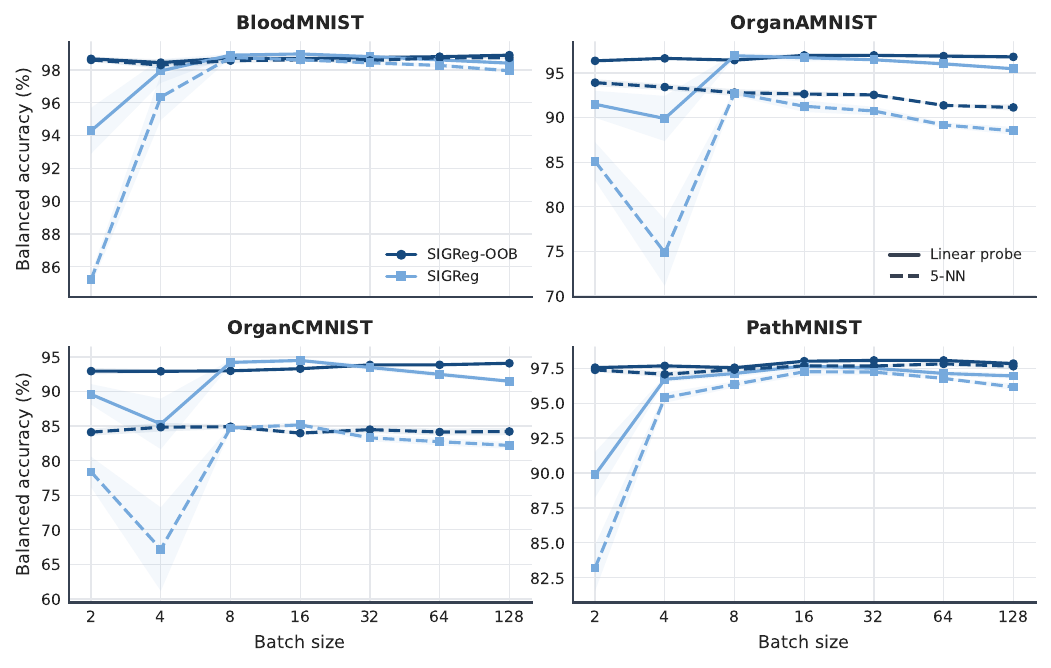}
  \caption{Per-dataset DINOv2 continued-pretraining performance on biomedical
  datasets for SIGReg and SIGReg-OOB. Solid lines show linear-probe accuracy
  and dashed lines show 5-NN accuracy. Each point averages DINOv2-S/B/L/G
  encoders; bands show standard errors across encoder sizes.}
  \label{fig:supp-continued-datasets}
\end{figure}

The per-dataset results (\Cref{fig:supp-continued-datasets}) support the main conclusion: at small batches,
OOB regularization is more likely to retain or improve the pretrained
representation than its in-batch counterpart. The benefit nevertheless
depends on the method, encoder, and target dataset, consistent with the central
claim being about access to population context rather than guaranteed
improvement for every pretrained representation.

\section{Computational measurements}
\label{app:compute}

\begin{figure}[h]
  \centering
  \includegraphics[width=0.95\textwidth]{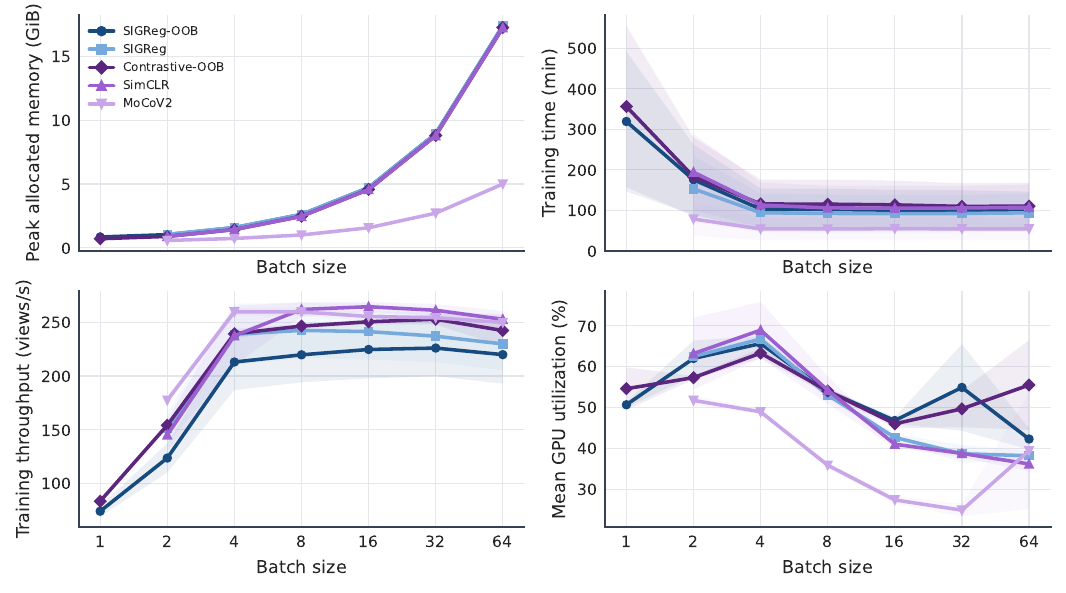}
  \caption{Detailed resource measurements across dataset sizes and batches.
  Lines average the datasets evaluated for each method, and bands show standard
  errors across datasets.}
  \label{fig:supp-resource}
\end{figure}

We benchmark each method on one NVIDIA L40S GPU for 100 epochs (\Cref{fig:supp-resource}). The five-method
benchmark spans BloodMNIST at 500, 1k, 2k, and 5k training examples and STL-10.
We evaluate batches 1--64 where each method is defined. Peak
allocated GPU memory is measured during training; wall time includes refresh
passes and anchor construction.

At a matched batch size, SIGReg-OOB and \contooob{} have essentially the same
peak allocated GPU memory as their direct SIGReg and SimCLR counterparts. They
add approximately 10\% and 4\% wall time, respectively, due to periodic cache
construction.

\section{Additional analyses}
\label{app:additional}

\subsection{Convergence over pretraining time}

\begin{figure}[h]
  \centering
  \includegraphics[width=0.88\textwidth]{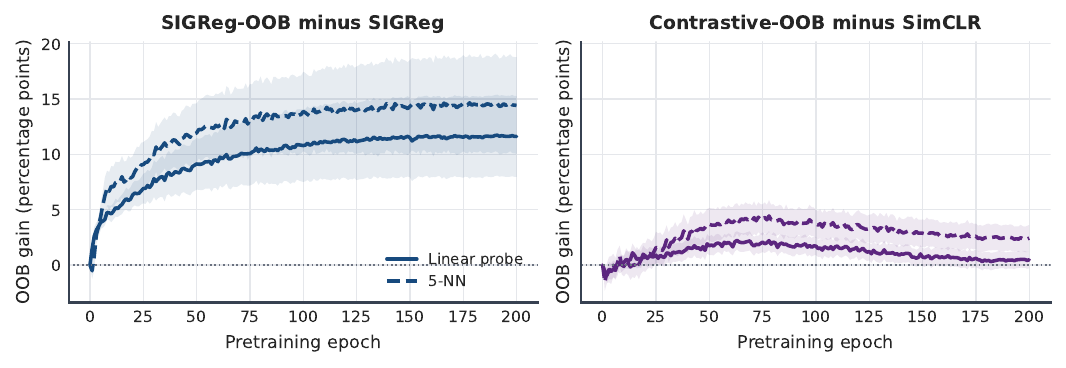}
  \caption{OOB improvement over the corresponding in-batch method during
  pretraining. Solid lines show linear-probe accuracy and dashed lines show
  5-NN accuracy. Lines average matched dataset--batch cells and bands show
  standard errors across those cells.}
  \label{fig:supp-convergence}
\end{figure}

We record frozen 5-NN and linear-probe performance after every epoch for
BloodMNIST-5k, STL-10, and ImageNet-100-5k. \Cref{fig:supp-convergence,tab:supp-convergence} shows
the difference in improvement between OOB and in-batch variants as optimization proceeds. The Gaussian OOB
advantage grows throughout training and the contrastive advantage emerges early
and remains positive on average through epoch 200.

\begin{table}[htbp]
  \centering
  \caption{Mean OOB gain in percentage points at selected epochs.}
  \label{tab:supp-convergence}
  \small
  \begin{tabular}{llrrrr}
\toprule
Comparison & Probe & Epoch 25 & Epoch 50 & Epoch 100 & Epoch 200 \\
\midrule
SIGReg-OOB $-$ SIGReg & 5-NN & +9.09 & +11.94 & +13.88 & +14.43 \\
SIGReg-OOB $-$ SIGReg & Linear & +6.91 & +9.14 & +10.80 & +11.62 \\
Contrastive-OOB $-$ SimCLR & 5-NN & +1.58 & +3.68 & +3.71 & +2.43 \\
Contrastive-OOB $-$ SimCLR & Linear & +0.89 & +1.73 & +1.71 & +0.47 \\
\bottomrule
\end{tabular}

\end{table}

\end{document}